%% file: main.tex
\documentclass[10pt]{article} 
\usepackage[preprint]{tmlr}

\input{math_commands.tex}

\usepackage{hyperref}
\usepackage{url}

\usepackage{amsmath, amssymb, amsthm}
\usepackage{booktabs}       
\usepackage{algorithm}
\usepackage[noend]{algpseudocode}
\newtheorem{theorem}{Theorem}[section]
\newtheorem{lemma}[theorem]{Lemma}
\newtheorem{definition}[theorem]{Definition}
\newcommand{\abs}[1]{\left\lvert #1 \right\rvert}
\newcommand{\F}{\mathcal{F}}

\newcommand{\norm}[1]{\left\lVert #1 \right\rVert}
\newcommand{\Z}{\mathbb{Z}}
\newcommand{\N}{\mathbb{N}}

\usepackage[pdftex]{graphicx} 
\newcommand\changes[1]{\textcolor{black}{#1}}

\title{Finite-sample Calibration Guarantees for\\ Binary Segmentation}

\author{\name Bruno Viti \email bruno.viti@uni-graz.at \\
      \addr IDea Lab - The Interdisciplinary Digital Lab at the University of Graz, Austria 
      \AND
      \name Martin Holler\thanks{MH is also a member of NAWI Graz \url{www.nawigraz.at} and BioTechMed Graz \url{https://biotechmedgraz.at/de/}} \email martin.holler@uni-graz.at \\
      \addr IDea Lab - The Interdisciplinary Digital Lab at the University of Graz, Austria\\
  }

\def\month{10}  
\def\year{2026} 
\def\openreview{\url{https://openreview.net/forum?id=fHHbzZZ2Na}} 

\begin{document}

\maketitle

\begin{abstract}
Machine learning–based segmentation models are frequently miscalibrated, producing for each pixel a label prediction whose associated confidence does not accurately reflect the true underlying probability. 
This limitation is particularly critical in high-stakes applications such as medical image segmentation. 
Although several calibration methods have been proposed, only a small subset of them provides theoretical guarantees on the calibration error, i.e., they do not typically ensure that, with high probability, the model exhibits a low calibration error. 
In this work, we develop a \textit{post-hoc} calibration method for which formal calibration guarantees can be established. 
To achieve this, we derive from standard concentration inequalities an improved theoretical bound that explicitly accounts for the spatial dependence inherent in binary segmentation outputs. 
We empirically demonstrate that the proposed calibrator satisfies the derived guarantee and outperforms existing \textit{post-hoc} calibration approaches.
\end{abstract}

\section{Introduction}
\label{introduction}
The calibration of a segmentation model is a fundamental task in machine learning that needs to be addressed in order to align the predicted confidence of a model with the true reliability of its prediction. 
It is essential to obtain reliable models for real-world applications. 
In the context of deep learning, the confidence of a model refers to the probability it assigns to its predicted output. 
Ideally, this confidence should faithfully reflect the true likelihood of correctness. 
However, modern neural networks tend to be overconfident, which means that their predicted confidence scores are poorly aligned with the actual accuracy \citep{guo2017calibration,ovadia2019can}. 
This misalignment becomes particularly crucial in critical-domain applications such as medical image segmentation, where erroneous predictions can have severe consequences. 
In such settings, a miscalibrated model may assign high confidence to incorrect segmentations, leading practitioners to place unwarranted trust in flawed outputs. 
In contrast, a well-calibrated model ensures that a predicted confidence of, say, 90\% corresponds in the long run to a true correctness rate of approximately 90\% \citep{dawid1982well}.

The calibration problem has been extensively studied, and existing approaches can broadly be categorized into three groups \citep{wang2023calibrating}. 
\emph{Post-hoc} calibration methods operate after training, adjusting the model's output probabilities without modifying its parameters. 
Classical techniques include Platt scaling \citep{platt1999probabilistic}, Temperature Scaling \citep{guo2017calibration}, Dirichlet calibration \citep{kull2019beyond}, Histogram Binning \citep{zadrozny2001obtaining}, Meta-cal \citep{ma2021meta}, and Isotonic Regression \citep{zadrozny2002transforming}. 
Although originally introduced for classification, these methods can be applied to segmentation by treating each pixel independently. 
Moreover, there are also spatially aware extensions, such as \citep{ding2021local, rousseau2025post, kuppers2022confidence}.
The second group, \emph{Train-time} calibration, has also been widely explored. 
Mixup \citep{zhang2017mixup}, CutMix \citep{yun2019cutmix}, and label smoothing \citep{szegedy2016rethinking,muller2019does} regularize confidence by exposing the model to soft or interpolated targets. 
This group also includes methods that modify the training loss to penalize miscalibration:
representative examples in the classification setting include Focal Loss~\citep{lin2017focal}, FLSD \citep{mukhoti2020calibrating}, MMCE \citep{kumar2018trainable}, and DECE \citep{bohdal2023metacalibration}, all of which encourage the model to reduce overconfidence while preserving discriminative performance. 
Several approaches have also been developed specifically for segmentation tasks, such as MDCA~\citep{hebbalaguppe2022stitch}, as well as methods tailored to the medical imaging domain \citep{barfoot2024average,barfoot2026average,murugesan2025neighbor,murugesan2023trust}.
The third group of calibration approaches are \emph{uncertainty estimation} methods, which includes Monte Carlo dropout techniques \citep{gal2016dropout,kendall2017uncertainties} and Deep Ensemble Networks \citep{lakshminarayanan2017simple, mehrtash2020confidence}. 
However, these methods, rather than recalibrating the model, primarily identify and highlight regions of high uncertainty.

In contrast to the works mentioned so far, which are purely empirical, methods that use calibration data to equip a pre-trained machine learning model with theoretical guarantees have seen a strong interest in the past years. 
Rooted in the work of \citet{vovk2005algorithmic}, several works leverage statistical learning theory, and in particular conformal prediction, to provide coverage guarantees for segmentation models \citep{brunekreef2024kandinsky,mossina2025conformal,viti2026consign,bereska2025sacp,teng2023predictive,liu2025spatial}. 
An interesting connection between conformal prediction and confidence calibration is developed by \citet{xi2025does}.
The authors first show that standard calibration approaches fail to improve conformal prediction, then propose a tailored calibration method designed to enhance prediction set efficiency.
These works, however, focus primarily on constructing reliable prediction sets, rather than recalibrating pre-trained models with provable guarantees on the calibration error itself.

A notable exception is the work of \citet{gupta2020distribution,gupta2021distribution}, which endows histogram-binning calibration with finite-sample statistical guarantees for classification tasks.
Following their notation, let the feature and label space be $\mathcal{X}$ and $\mathcal{Y}=\{0,1\}$.
We have at our disposal a pre-trained predictor $f : \mathcal{X} \rightarrow [0,1]$, where $f(X)$ is the (uncalibrated) predicted probability that $Y=1$,
and a calibration dataset $\mathcal{D}_{cal} = \{(X^{(i)}, Y^{(i)})\}_{i=1}^{N_{cal}}$. 
The pre-trained model $f$ belongs to the space of all measurable functions $\mathcal{F}$ from $\mathcal{X}\to [0,1]$. 
We would like to construct a calibrator $H:\mathcal{F}\times \mathcal{D}_{{cal}}\to \mathcal{F}$, that recalibrates $f\in\mathcal{F}$ given the dataset $\mathcal{D}_{{cal}}$. 
We denote the calibrated predictor as $h:=H(f, \mathcal{D}_{{cal}})$. A predictor is perfectly calibrated if $\E[Y|h(X)]=h(X)$, which, in the binary case, corresponds to $\mathbb{P}(Y=1\mid h(X))=h(X)$. 
While the latter condition is impossible to guarantee in a finite-sample setting, we would like to have a guarantee on the calibration error using the following notions:
\begin{definition}
    (Marginal calibration). Let $\varepsilon>0$ and $\alpha\in[0,1]$. A calibrator $H$ is said to be $(\varepsilon,\alpha)$-marginally calibrated if for every predictor $f\in\mathcal{F}$ and distribution $\mathbb{P}$ over $\mathcal{X}\times [0,1]$
    \[\mathbb{P}(|\E[Y\mid h(X)]-h(X)|\leq\varepsilon)\geq 1-\alpha.\]
    \label{def1}
\end{definition}
\begin{definition}
    (Conditional calibration). Let $\varepsilon>0$ and $\alpha\in[0,1]$. A calibrator $H$ is said to be $(\varepsilon,\alpha)$-conditionally calibrated if for every predictor $f\in\mathcal{F}$ and distribution $\mathbb{P}$ over $\mathcal{X}\times [0,1]$
    \[\mathbb{P}(\forall r\in Range(h):|\E[Y\mid h(X)=r]-r|\leq\varepsilon)\geq 1-\alpha.\]
    \label{def2}
\end{definition}
The fundamental distinction between the two definitions is that in the marginal case the probability is taken over both $X$ and $\mathcal{D}_{cal}$, meaning that the probability holds only on average over $h(X)$. 
Conditional calibration instead, which implies marginal, requires the bound to hold uniformly in $r$. 

The seminal work of \citet{gupta2020distribution,gupta2021distribution} remains confined to the classification setting. 
Our contribution builds directly on this line of work and extends it beyond classification.
To the best of our knowledge, no existing work provides marginal and conditional guarantees in the segmentation setting. 
One reason is that standard approaches require large amounts of independent and exchangeable calibration data — an assumption that breaks down in segmentation, where pixel-wise predictions within an image are spatially correlated. 
This paper fills this gap by leveraging concentration inequalities, specifically adapted to account for spatial pixel correlation in our setting, to derive calibration guarantees for segmentation models.

The main theoretical tool used in \citep{gupta2020distribution,gupta2021distribution} to bound the calibration error is Hoeffding's inequality \citep{hoeff}, which provides concentration for sums of bounded $i.i.d.$ random variables. 
In segmentation, however, the predictor outputs a probability for each pixel, and the spatial correlation between pixels violates the $i.i.d.$ assumption, making such inequalities inapplicable. 
While extensions to the non-$i.i.d.$ setting exist — including martingale-based results such as \citet{azuma1967weighted} and \citet{freedman1975tail} inequalities, and concentration under $\beta$-mixing conditions \citep{yu1994rates, mohri2010stability} — these do not capture the structured spatial dependencies inherent to segmentation outputs.

Image segmentation outputs can naturally be modeled as random fields: each pixel label 
depends not only on its own features but also on the label of its spatial neighbors, 
inducing dependencies across the output map that are well captured by random field 
models such as Markov Random Fields \citep{li2009markov, zoltan2012markov}. Bounding the calibration error in this setting therefore requires concentration inequalities that explicitly account for such spatial dependencies.
Early approaches to random fields relied on information-theoretic tools: 
\citet{marton1998,marton2003} and \citet{samson2000} established concentration results 
for $\Phi$-mixing processes using transportation and Pinsker inequalities, albeit restricted to 
exponential bounds. A significant step forward was achieved by \citet{chazottes2007}, who 
proposed a coupling-based framework for random fields indexed by $\mathbb{Z}^d$ that yields concentration inequalities depending on the 
strength of spatial dependence. A parallel martingale-based approach was developed in \citep{kontorovich2008concentration}, 
yielding concentration bounds for Markov and hidden Markov chains in terms of mixing 
coefficients. More recently, \citet{garnier2022} established a Hoeffding-type inequality 
specifically for non-causal random fields. Finally, 
\citet{chazottes2026finitary} provides sharp Gaussian concentration criteria for Gibbs measures 
and Markov random fields on $\mathbb{Z}^d$.
\paragraph{Our contribution}
In this work, we address the problem of calibration for binary segmentation models.
We first derive an improved concentration inequality tailored to variables exhibiting the structured spatial dependence of binary segmentation outputs.
Building upon the framework of \citep{chazottes2007}, we model the spatial relationship between pixels and obtain a tighter theoretical bound. 
We then apply the bound to the calibration setting: we construct a calibrator $H$ following the idea of \citet{gupta2021distribution}, and establish that the calibrator satisfies the theoretical guarantees provided by our bound. 
This result formally justifies its application in the segmentation setting. A key insight is that treating each individual pixel prediction as a data point vastly increases the effective size of the calibration set. 
As a result, our method achieves tight statistical guarantees with a comparably small number of calibration images — even with only 50 images — making it practical for real-world applications where calibration data is scarce. 
Beyond providing theoretical guarantees, we also show that the calibrator $H$
even practically outperforms other \textit{post-hoc} calibration approaches that do not come with such guarantees.

The outline of the paper is the following: In Section \ref{th_bnd} we derive the improved theoretical bound and in Section \ref{cal} we show how the bound can be exploited to obtain statistical guarantees for calibration errors in the segmentation setting. 
Finally, in Section \ref{D_est} we show how to estimate the bound and in Section~\ref{experiments} we empirically show the validity of our approach.


\section{Concentration Inequality under Spatial Dependence}
\label{th_bnd}
For this section we follow the notation of \citet{chazottes2007}. Let $A$ be a finite set, $g:A^n\xrightarrow{} \mathbb{R}$ a function of $n$ variables, and $\sigma=(\sigma_1,\dots,\sigma_i,\dots)$ an element of $A^{\mathbb{N}}$. As in \citep{chazottes2007}, we consider, with an abuse of notation, $g$ as a function on $A^\mathbb{N}$, which does not depend on $\sigma_k$ for $k>n$. 
We make use of two quantities, both defined later in this section: the coupling matrix $\bar{D}$, which encodes the dependence structure among the $n$ variables and quantifies how well the distribution of each variable is determined by the others; and the sensitivity $\delta g$, which measures how much $g$ can change when a single element of $\sigma$ is perturbed. 
With this notation, \citet{chazottes2007} derive concentration inequalities of the form:\[
  \mathbb{P}(\abs{g-\E[g]}\geq \varepsilon)
  \leq 2\exp\!\left(\frac{-2\varepsilon^2}{\|\bar D\|^2_{\ell^{2}(\mathbb{N})}\,\|\delta g\|^2_{{\ell^{2}(\mathbb{N})}}}\right).
\]  
We are going to improve the bound through two assumptions that hold true in our setting, namely: ${A=\{0,1\}}$ is a Bernoulli alphabet and $g$ is the average of Bernoulli variables. 
\subsection{Bernoulli alphabet} 
To establish the concentration inequality, we use a martingale argument which decomposes the quantity of interest $g-\mathbb{E}[g]$ as the sum of martingale differences \cite{mcdiarmid1989method}.
For $i=1,\dots,n$ let $\mathcal{F}_i$ be the sigma-field generated by the random variables $\sigma_1,\dots,\sigma_i$. 
Then, following \citet[Section~2.1]{chazottes2007}, we write \begin{equation}
g(\sigma)-\E[g]=\sum^n_{i=1}V_i(\sigma),\qquad \text{where}\quad V_i(\sigma):=\mathbb{E}[g|\mathcal{F}_i](\sigma)-\mathbb{E}[g|\mathcal{F}_{i-1}](\sigma).
\label{eq:martingaledecomp}
\end{equation} 
Further, we write $\sigma_{<i}$ for the values at sites ordered before $i$, and define
\[
  p_i(\sigma) := \mathbb{P}(\sigma_i = 1 \mid \sigma_{<i}), \qquad
  \Delta_i(\sigma) := \E[g \mid \sigma_{<i},\, \sigma_i = 1] - \E[g \mid \sigma_{<i},\, \sigma_i = 0].
\]
For a Bernoulli alphabet $A=\{0,1\}$, a computation using the law of total expectation (see \ref{A:Vi}) yields:
\begin{equation}\label{eq:Vi}
  {V_i(\sigma) = (\sigma_i - p_i)\,\Delta_i.}
\end{equation}
To obtain the concentration inequality, we have to bound the moment generating function (MGF) of each variable $V_i$. 
However, \citet[Lemma~1]{chazottes2007} uses only the range of $V_i$, namely $\Delta_i$, to obtain such a bound. 
We, instead, make use of the fact that $V_i$ is a scaled centered Bernoulli with conditional variance $\Var(V_i\mid\mathcal{F}_{i-1})=p_i(1-p_i)\Delta_i^2$, 
and exploit the terms $p_i(1-p_i)$ to provide a tighter MGF bound than the range-only estimate.
\changes{This choice does not come from a formal comparison between the Bennett and the range-only bounds, but is rather qualitative, driven by the intuition, later supported by the experiments, that the conditional variance $p_i(1-p_i)$ is typically small and thus worth retaining in the bound.}
To do that, we apply Bennett's inequality \citep[Theorem~2.9]{Massart} and prove it in Appendix~\ref{pr:lemmabenent}:
\begin{lemma}[Bennett bound]\label{lem:bennett}
Let $V_i=(\sigma_i-p_i)\Delta_i$ and $\Delta_i\geq0$,\ then $\forall \lambda > 0$ :
\begin{equation}\label{eq:bennett-cond}
  \log\E\bigl[e^{\lambda V_i}\mid\F_{i-1}\bigr]
  \leq p_i(1-p_i)\,\phi(\lambda\Delta_i),
  \qquad \text{where} \quad \phi(u) := e^u - u - 1.
\end{equation}
\end{lemma}
\subsection{The coupling matrix $\bar D$ and the function $g$}
\label{subsec:Dg}
Now that we have bounded the MGF of $V_i$ by $p_i(1-p_i)\,\phi(\lambda\Delta_i)$, we need to bound the increment $\Delta_i$. 
To this aim, for a given configuration $\sigma$, we define the coupling matrix $D^\sigma$ as
\[
D^\sigma_{i,j} := \mathbb{P}^{\sigma_{<i}}_{i,0,1}\!\left( \sigma^{(1)}_j \neq \sigma^{(2)}_j \right),
\]
where $\mathbb{P}^{\sigma_{<i}}_{i,0,1}$ denotes the maximal coupling \citep{lindvall2002lectures} of the conditional
distributions $\mathbb{P}(\cdot|\sigma_{<i},\sigma_i=1)$ and $\mathbb{P}(\cdot|\sigma_{<i},\sigma_i=0)$. Intuitively, $D^\sigma_{i,j}$
measures the influence of site $i$ on site $j$: it is the probability that two coupled copies ($\sigma^{(1)}$ and $\sigma^{(2)}$) of the process, agreeing on the past $\sigma_{<i}$, but initialized with opposite values at site $i$, still disagree at site $j$. 
Notice that for $j<i$ we have $D^\sigma_{i,j}=0$ and that, in case of independent random variables, the matrix $D^\sigma$ would be the identity matrix. 
With $D^\sigma$ in hand, we can now bound $\Delta_i$ via the variation of $g$.
In our setting, $g$ is the empirical mean of Bernoulli variables, and its variation at site $i$ is
\begin{equation} 
\delta_i g = \sup_{\substack{\sigma,\,\sigma'\\ \sigma_j=\sigma'_j,\;\forall j\neq i}}
    |g(\sigma)-g(\sigma')| = \frac{1}{n},\quad g(\sigma)= \frac{1}{n}\sum_{i=1}^n \sigma_i.
    \label{eq:gdef}  
\end{equation}

The last step is using \citet[Eq.~5]{chazottes2007} to bound the increment by $\Delta_i(\sigma)\leq(D^\sigma\delta g)_i$. 
In our case, since $\delta_j g = 1/n$ for all $j$, and defining $\bar{D}_{i,j} := \sup_\sigma D^\sigma_{i,j}$ to remove the dependence on~$\sigma$, this simplifies to (see Appendix~\ref{A:delta}):
\begin{equation}\label{eq:deltabound}
\Delta_i(\sigma) \leq \frac{1}{n}\sum^{n}_{\changes{j\geq i}} \bar{D}_{i,j}:=\frac{R_i}{n}, 
\end{equation}
where $R_i$ denotes the $i$-th row sum of~$\bar{D}$, i.e., the cumulative forward influence of element $i$ on all subsequent elements.
A more detailed discussion and practical estimation of $\bar{D}$ is provided in Section~\ref{D_est}.

\subsection{Main Theorem}
We now state the concentration inequality and prove it in Appendix~\ref{pr:proofofthmimproved}.
\begin{theorem}[Concentration for Bernoulli mean]%
\label{thm:improved}
Let $g = n^{-1}\sum^n_{i=1}\sigma_i$.
Define the coupling matrix $\bar D$ as in Section~\ref{subsec:Dg},
$R_i := \sum^n_{\changes{j\geq i}} \bar D_{i,j}$, and
\changes{$\bar q_i=\sup_{\sigma_{<i}}p_i(1-p_i)$} with $p_i := \mathbb{P}(\sigma_i=1\mid\sigma_{<i})$.
Set
\begin{equation}\label{eq:BC}
  \changes{S \;:=\; \frac{1}{n^2}\sum^n_{i=1} \bar q_i\,R_i^2,}
  \qquad
  R_{\max} := \max_{i=1,\dots,n} R_i, \quad c \;:=\; \frac{R_{\max}}{3n}.
\end{equation}
Then for all $\varepsilon > 0$:
\begin{equation}\label{eq:main}
  \boxed{\mathbb{P}(\abs{g - \E[g]} \geq \varepsilon)
  \leq
  2\exp\left(-\frac{S}{c^2}\eta\Big(\frac{c\varepsilon}{S}\Big)\right)}\qquad with \quad \eta(x)=1+x-\sqrt{1+2x}.
\end{equation}
\end{theorem}
Notice that the notation $\sigma_{<i}$ relies on $\mathbb{N}$ having a natural total order.
For variables indexed in $\mathbb{Z}^2$, one fixes an arbitrary enumeration $\Gamma:\mathbb{Z}^2\to\mathbb{N}$ and writes $\sigma_{<x}:=\{\sigma_y:\Gamma(y)<\Gamma(x)\}$.
The full extension to $\mathbb{Z}^2$ is given in Appendix~\ref{subsec:extensionZ2} and involves mainly this notational variation.


\section{Calibration Guarantees}
\label{cal}
We are interested in obtaining calibration error bounds in the segmentation setting where the $i.i.d.$ assumption does not hold due to the spatial dependence between pixels. 
Let ${x \in \Omega := \{1,\dots,H\} \times \{1,\dots,W\}}$ denote pixel positions. 
Let $\mathcal{X}=[0,1]^{\Omega}$ and $\mathcal{Y}=\{0,1\}^{\Omega}$ be the feature and label spaces of images and their segmentation masks, respectively. 
Let $f:\mathcal{X}\to [0,1]^{\Omega}$ be a pre-trained predictor, where $f(X)_{x}$ predicts the probability of $Y_{x}=1$.
We seek a calibrator $H$ inducing a calibrated predictor $h$, which maps each pixel-wise confidence $f(X)_{x}$ to a revised confidence score. 
We are going to use a binning approach, first introduced in \citep{zadrozny2001obtaining} and more recently used in \citep{gupta2020distribution,gupta2021distribution}. 
Let $\mathcal{D}=\{X^{(i)},Y^{(i)}\}^{N}_{i=1}$ be a dataset featuring images and binary segmentations.
We partition the dataset into two subsets $\mathcal{D}_{\mathrm{bin}}$ and $\mathcal{D}_{\mathrm{cal}}$, using the first one to compute the binning scheme and the latter to construct the calibrator $H$.
\subsection{Binning scheme}
The first step is to create a binning scheme that provides a partition $\{\Omega_b\}^B_{b=1}$ of the pixel positions. 
Following the definition of \citet{gupta2020distribution}, a binning scheme $\mathcal{B}=\{I_b\}^B_{b=1}$ is any partition of $[0,1]$ into $B$ non-overlapping intervals $I_1,\dots,I_B$ such that $\bigcup^B_{b=1}I_b=[0,1]$ and $I_b\cap I_{b'}=\emptyset$ for $b\neq b'$. 
Therefore, given $\mathcal{B}$ and $f$, the partition for a generic image $i$ is defined as \[\Omega^{(i)}_b=\{x\in\Omega:\ f(X^{(i)})_x\in I_b \}.\]
A naive way to construct the binning scheme is to choose fixed-width bins such that $I_j=[\frac{j-1}{B},\frac{j}{B})$, $j=1,\dots B-1$ and $I_B=[\frac{B-1}{B},1]$. 
This binning scheme requires no held-out data, but tends to produce highly imbalanced bins, since overconfident predictors typically assign extreme probabilities (close to 0 or 1) to most pixels. 
Therefore, we empirically find a uniform-mass binning scheme through quantiles \citep{gupta2021distribution}. 
The bins are defined as \[I_1=[0,\hat q_1),\ I_j=[\hat q_{j-1},\hat q_{j}],\ j=2,\dots,B-1,\ I_B=(\hat q_{B-1},1],\quad \hat{q}_j = \mathcal{Q}_{\frac{j}{B}}(\{f(X^{(i)})_x\}^{\abs{\mathcal{D}_{bin}}}_{i=1,x\in\Omega}).\]
where $\hat q_j$ is the $(j/B)$-th quantile, empirically estimated using the images in $\mathcal{D}_{bin}$.
\subsection{Calibrator}
\label{subsec:calibrator}
Now that we have a binning scheme $\mathcal{B}$, we partition the pixels of the $N_{cal}$ images of $\mathcal{D}_{cal}$ using $\mathcal{B}$, and we define the calibrated predictor $h=H(f,\mathcal{D}_{cal})$ as:
\begin{equation}h(X)_x=\sum^B_{b=1}\mathbf{1}_{\{f(X)_x\in I_b\}}\hat \pi_b,\quad \text{where}\quad \hat \pi_b=\frac{1}{N_b}\sum^{N_{cal}}_{i=1}\sum_{x_b\in\Omega^{(i)}_b}Y_{x_b}^{(i)},\ N_b=\sum^{N_{cal}}_{i=1}|\Omega^{(i)}_b|.\label{definition_estimator}\end{equation}
As an example, suppose that on the calibration dataset bin $b$ has an average raw confidence (predicted by $f$) of $0.8$, but only $70\%$ of those predictions have ground-truth equal to $1$. 
Then, for a new test point $X_{test}$, the calibrator $H$ replaces the raw confidence $f(X_{test})_x$ with the empirical label frequency $\hat\pi_b$ observed in bin $b$ on the calibration set, i.e., $h(X_{test})_x=\hat\pi_b=0.7$ if $f(X_{test})_x\in I_b$.
We note that $\hat\pi_b$ is to be understood as conditioned on the binning scheme $\mathcal{B}$, so that $N_b$ is treated as a constant and the only source of randomness are the Bernoulli labels. 
The calibration procedure is summarized in Algorithm \ref{tab:alg}.
\subsection{Calibration Guarantees Theorems}
\label{subsec:randomcalibrator}
We now use the theory derived in Section \ref{th_bnd} to bound the calibration error with the following theorems, which are all proved in Appendix~\ref{A:cal}.
Before stating the calibration theorems, observe that for a fixed bin $b$, $\hat\pi_b$ is the empirical mean of $N_b$ dependent Bernoulli variables. 
Thus, $\hat\pi_b$ plays the role of $g(\sigma)$, with $\sigma$ being the labels of pixels falling in bin $b$, i.e., $\sigma=\{Y_{x_b}^{(i)}\}^{N_{cal}}_{i=1,  x_b\in\Omega^{(i)}_b}$.

\begin{theorem}[Per-bin concentration]
\label{ourbound1}
    Let ${\mathcal{B}=\{I_b\}^B_{b=1}}$ be a fixed binning scheme. 
    Then, given a calibration set $\{(X^{(i)},Y^{(i)})\}^{N_{cal}}_{i=1}$, let $\{\Omega^{(i)}_b\}_{b=1}^B, i=1,...,N_{cal}$ be the corresponding partitions and define the observed frequency $\hat{\pi}_b$ in bin $b$ as in \eqref{definition_estimator}. 
    Denote as $E_{\mathcal{B}}$ a realization of the binning event that fixes the calibration labels $Y_{x_b}^{(i)}$ in the corresponding bins, and fix a bin $b$.
    Let $\bar D^{(i)}$ be the coupling matrix (as defined in Section~\ref{subsec:Dg}) of image $i$ conditioned on the binning event $E_{\mathcal{B}}$ (i.e. the collection $\{Y^{(i)}_{x_b}\}_{x_b\in\Omega^{(i)}_b}$), and define
    \[
      R^{(i)}_{x_b} = \sum_{\changes{y_b \geq x_b}} \bar D^{(i)}_{x_b,y_b}, \qquad
      p^{(i)}_{x_b} = \mathbb{P}\bigl(Y^{(i)}_{x_b}=1\mid Y^{(i)}_{<x_b},E_{\mathcal{B}}\bigr), \qquad
      \changes{\bar{q}^{(i)}_{x_b} = \sup_{Y^{(i)}_{<x_b}} p^{(i)}_{x_b}(1-p^{(i)}_{x_b})}
    \]
    Set
    \[
      S_b = \frac{1}{N_b^2}\sum^{N_{cal}}_{i=1}\sum_{x_b\in\Omega^{(i)}_b} \changes{\bar{q}^{(i)}_{x_b}}\,(R^{(i)}_{x_b})^2, \quad
      R_{\max,b} = \max_{i=1,\dots,N_{cal}}\max_{x\in\Omega^{(i)}_b} R^{(i)}_{x_b}, \quad
      c_b = \frac{R_{\max,b}}{3N_b}.
    \]
    Then for all $\varepsilon > 0$:
    \[\mathbb{P}(\abs{\hat\pi_b-\E[\hat\pi_b\mid E_{\mathcal{B}}]}\geq \varepsilon)\leq 2\exp\!\left(-\frac{S_b}{c_b^2}\,\eta\!\left(\frac{c_b\, \varepsilon}{S_b}\right)\right),\quad with\quad  \eta(x)=1+x-\sqrt{1+2x}.\]
\end{theorem}
\begin{theorem}[Uniform bound over bins]
\label{ourbound2}
    Let the assumptions of Theorem~\ref{ourbound1} hold. 
    Then:
    \[\mathbb{P}(\exists\, b\in[B]:\abs{\hat\pi_b-\E[\hat\pi_b\mid E_{\mathcal{B}}]}\geq \varepsilon)\leq 2\sum^B_{b=1}\exp\!\left(-\frac{S_b}{c_{{b}}^2}\,\eta\!\left(\frac{c_{b} \varepsilon}{S_{b}}\right)\right),\quad with\quad  \eta(x)=1+x-\sqrt{1+2x}.\]
  \end{theorem}

In the segmentation setting we apply Definitions \ref{def1}-\ref{def2} pixel-wise, replacing $Y$ and $h(X)$ by the per-pixel quantities $Y_x$ and $h(X)_x$, so that the guarantee is required to hold at an arbitrary fixed pixel $x \in \Omega$. 
This is the natural scope for our binning-based calibrator, which acts on each pixel independently. 
However, to connect Theorems~\ref{ourbound1}–\ref{ourbound2} to this notion, observe that for a variable $Y_x$, conditioning on the binning event $E_{\mathcal{B}}=\{f(X)_x\in I_b\}$ does not in general coincide with conditioning on $h(X)_x$: 
the two agree only when $\hat\pi_{b'}\neq \hat\pi_{b}$ for every $b'\neq b$. 
Following \citet{gupta2021distribution}, we resolve this by adding an arbitrarily small uniform perturbation $\delta$ to $h(X)_x$, e.g. $\delta=10^{-20}$, to obtain a randomized version $\tilde{h}(X)_x$. The randomized version is practically indistinguishable from the original calibrated predictor $h(X)_x$ but makes the binning scheme recoverable from $\tilde{h}(X)_x$. 
The explicit construction is deferred to the proof of Theorem~\ref{marandcon_cal} and the randomized calibration procedure is summarized in Algorithm \ref{tab:alg}.
We now state the new calibration guarantees:
\begin{theorem}[Calibration guarantees]
    Let the assumptions of Theorem~\ref{ourbound1} hold. Let $\tilde{H}$ be the randomized calibrator parametrised by $\delta$, such that $\tilde{h}=\tilde{H}(f,\mathcal{D}_{cal})$. 
    Assume moreover that ${Y_x\mid \tilde{h}(X)_x\perp\!\!\!\perp \tilde{h}(X)_y,\ y\neq x}$, and that, by tanslation invariance, pixels share the same conditional mean ${\E[Y_{x}\mid \tilde{h}(X)_{x}]}={\E[Y_{y}\mid \tilde{h}(X)_{y}]},\ \forall x,y$.
    Define
    \[\alpha_{1,b}={2\exp\!\left(-\frac{S_b}{c_b^2}\,\eta\!\left(\frac{c_b\, (\varepsilon-\delta)}{S_b}\right)\right)},\ \alpha_2=2\sum^B_{b=1}\exp\!\left(-\frac{S_b}{c_{{b}}^2}\,\eta\!\left(\frac{c_{b} (\varepsilon-\delta)}{S_{b}}\right)\right).\]
    Then, $\tilde{H}$ is ${(\varepsilon,\alpha_{1,b})}$-marginally calibrated for every $b=1,\dots,B$ and every $x_b\in\Omega_b$:
    \[\mathbb{P}\bigg(\abs{\E[Y_{x_b}\mid \tilde{h}(X)_{x_b}]-\tilde{h}(X)_{{x_b}}}\leq \varepsilon\bigg)\geq1-\alpha_{1,b}.\]
    Moreover $\tilde{H}$ is $(\varepsilon,\alpha_2)$-conditionally calibrated for every $x\in\Omega$:
    \[\mathbb{P}\bigg(\forall\, r\in Range(\tilde{h}):\abs{\E[Y_{{x}}\mid \tilde{h}(X)_x=r]-r}\leq \varepsilon\bigg)\geq1-\alpha_2.\]
\label{marandcon_cal}
\end{theorem}
The assumption $Y_x\mid \tilde{h}(X)_x\perp\!\!\!\perp \tilde{h}(X)_y,\ y\neq x$ states that, once the bin assignment at pixel $x$ is known, the bin assignments at all other pixels provide no additional information about the label $Y_x$.
Notice that the pre-trained network $f$ is itself designed to aggregate neighborhood context into the prediction $f(X)_x$, so the bin assignment at $x$ already inherits this spatial information.
Crucially, the assumption does not make the problem trivial---the Bernoulli labels $Y_x$ remain dependent across pixels, which is why the spatial concentration inequality is needed for our argument.
\changes{We acknowledge, however, that this conditional independence assumption may be too strong in some settings, and we discuss this aspect further in Appendix \ref{app:addexp}.}
 
\begin{algorithm}
 \caption{Calibration algorithm}
\label{tab:alg}
\begin{algorithmic}[1]
\Statex\textbf{Input:} number of bins $B$, pre-trained predictor $f$, dataset $\{(X^{(i)},Y^{(i)})\}_{i=1}^{N}$, randomization parameter $\delta\ge0$ (set $\delta=0$ for calibrator of Subsection  \ref{subsec:calibrator}, $\delta>0$ for the randomized version  \ref{subsec:randomcalibrator})
\Statex\textbf{Output:} calibrated predictor $\tilde{h}$
\State Partition the dataset into $\mathcal{D}_{\mathrm{bin}}$ and $\mathcal{D}_{\mathrm{cal}}=\{(X^{(i)},Y^{(i)})\}^{N_{cal}}_{i=1}$
\Statex \textit{// Binning scheme: uniform-mass binning via quantiles on $\mathcal{D}_{\mathrm{bin}}$}
\State Collect the pool of predicted confidences $\mathcal{P}\gets\{f(X^{(i)})_x:\ i\in\mathcal{D}_{\mathrm{bin}},\ x\in\Omega\}$
\For{$j=1,\dots,B-1$}
    \State $\hat q_j \gets \mathcal{Q}_{j/B}(\mathcal{P})$ \Comment{empirical $(j/B)$-th quantile}
\EndFor
\State $I_1\gets[0,\hat q_1),\ \ I_j\gets[\hat q_{j-1},\hat q_j]$ for $j=2,\dots,B-1,\ \ I_B\gets(\hat q_{B-1},1]$
\Statex \textit{// Calibrator: empirical label frequency per bin on $\mathcal{D}_{\mathrm{cal}}$}
\For{$i=1,\dots,N_{cal}$ and $b=1,\dots,B$}
    \State $\Omega^{(i)}_b\gets\{x\in\Omega:\ f(X^{(i)})_x\in I_b\}$ \Comment{assign pixels to bins}
\EndFor
\For{$b=1,\dots,B$}
    \State $N_b\gets\sum_{i=1}^{N_{cal}}|\Omega^{(i)}_b|$
    \State $\hat\pi_b\gets\dfrac{1}{N_b}\sum_{i=1}^{N_{cal}}\sum_{x_b\in\Omega^{(i)}_b}Y^{(i)}_{x_b}$ \Comment{empirical positive-class frequency}
    \State $U_b\sim\mathcal{U}[-1,1]$ \Comment{used only when $\delta>0$}
\EndFor
\State \Return $\tilde{h}(X)_x=\displaystyle\sum_{b=1}^{B}\mathbf{1}_{\{f(X)_x\in I_b\}}\,(\hat\pi_b+\delta U_b)$
\end{algorithmic}
\end{algorithm}


\section{Estimation of Theoretical Bound}
\label{D_est}
Since the bound is not analytically tractable, due to the matrix $\bar D$, we estimate it empirically. While this does not yield a closed-form guarantee, it provides a principled, quantitative basis for evaluating the binning method and guiding its application in the segmentation setting.
Let $\Omega \subset \mathbb{Z}^2$ be a finite rectangular image domain with pixel alphabet $A = \{0, 1\}$, so configurations $\sigma$ are elements of $\{0,1\}^\Omega$.
We fix an arbitrary ordering $\Gamma : \Omega \to \{1, \dots, |\Omega|\}$
of the pixels, and we write $\sigma_{<x}$ for the configuration on $\{y \in \Omega : \Gamma(y) < \Gamma(x)\}$.
Following~\citep[Eq.~4]{chazottes2007}, and using the total-variation formulation of~\citep[Eq.~2.1]{samson2000}, we define the matrix $D^\sigma$ with entries indexed by pairs
$(x, y) \in \Omega \times \Omega$ with $\Gamma(y) > \Gamma(x)$ as
\begin{equation}\label{eq:Dsigma}
  D^\sigma_{x,y}
  =
  \left\|
    \mathbb{P}\!\left(\sigma_y \mid \sigma_{<x},\sigma_x=1\right)
    -
    \mathbb{P}\!\left(\sigma_y \mid \sigma_{<x},\sigma_x=0\right)
  \right\|_{\mathrm{TV}},
\end{equation}
and $D^\sigma_{x,y} = 0$ for all other pairs.
This formulation is obtained by choosing the Hamming metric in the definition of $D^\sigma$ \citep[Eq.~4]{chazottes2007} and is equivalent to the coupling matrices defined in \citet[Eq.~1.2]{kontorovich2008concentration} and \citet[Eq.~2.1]{samson2000}.
In our application, $\sigma$ is the vector of labels $Y_x$ restricted to the pixels $x_b$ that are contained in a fixed partition $\Omega_b$, i.e., $\sigma=\{Y_{x_b}\}_{x_b\in\Omega_b}$.
Moreover, as detailed in Section~\ref{cal}, the variables $Y_x$ are conditioned on the binning event $E_{\mathcal{B}}$ and therefore on $h(X)_x$.
Adapting~\eqref{eq:Dsigma} to this conditional process and taking the 'worst' configuration to remove the $\sigma$ dependence, the coupling matrix that enters Theorem~\ref{ourbound2} becomes
\begin{equation}
  \label{eq:Dsigma_bound}
  \bar D_{x_b,y_b}
  =\sup_{\sigma_{<x_b}}
  \abs{
    \mathbb{P}\!\left(\sigma_{y_b}=1 \mid \sigma_{<x_b},\,\sigma_{x_b}=1,\,h(X)_{y_b}\right)
    -\mathbb{P}\!\left(\sigma_{y_b}=1 \mid \sigma_{<x_b},\,\sigma_{x_b}=0,\,h(X)_{y_b}\right)
  },
\end{equation}
since for a binary alphabet the total-variation distance equals the absolute difference of the Bernoulli means.

\begin{figure}
  \centering
    \includegraphics[width=0.3\linewidth]{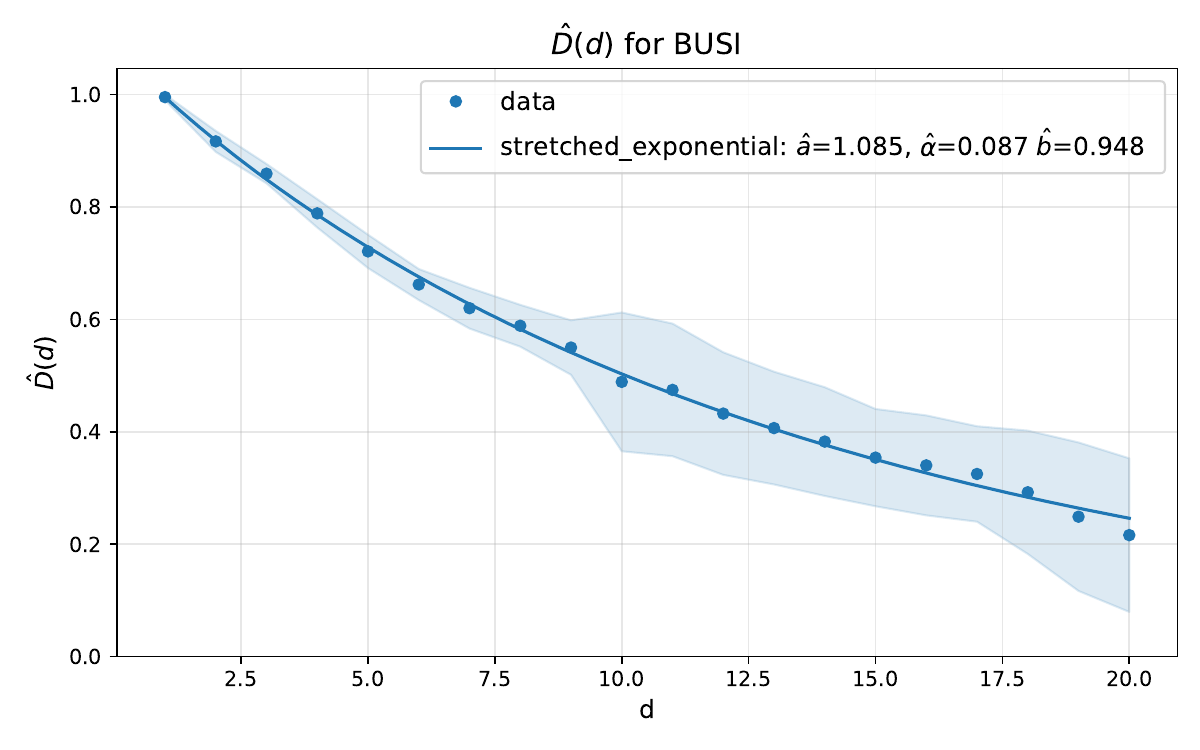}
    \includegraphics[width=0.3\linewidth]{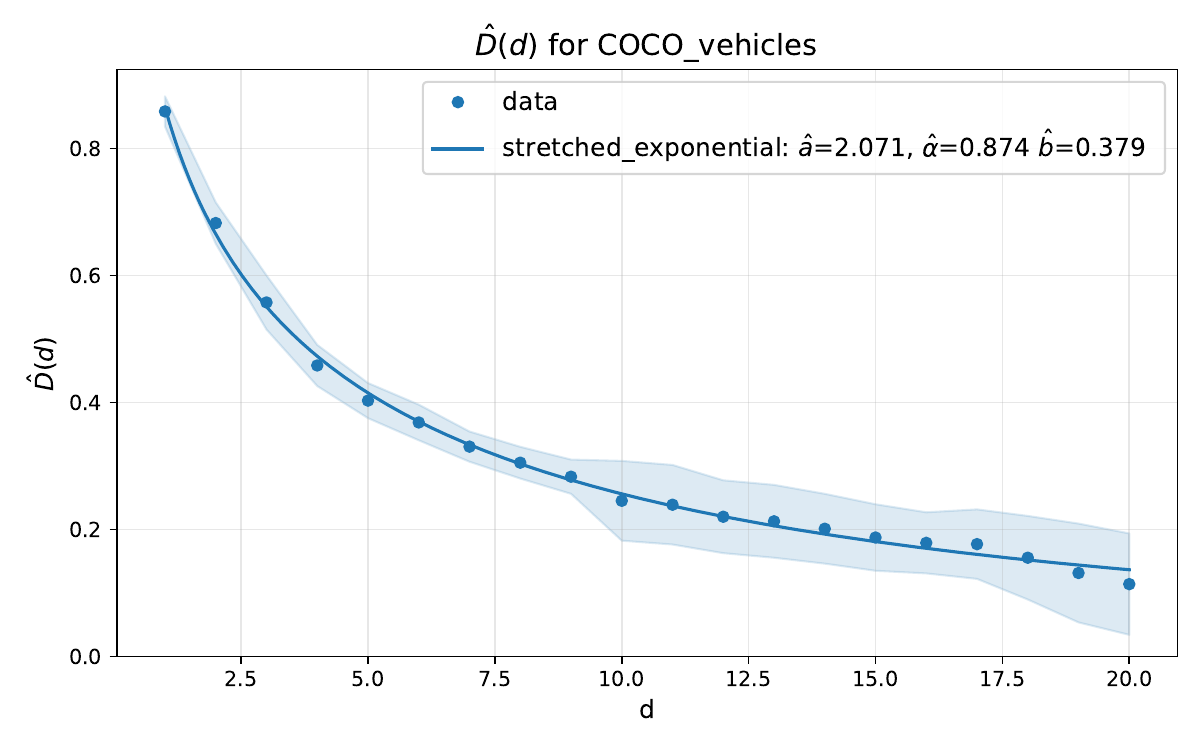}
    \includegraphics[width=0.3\linewidth]{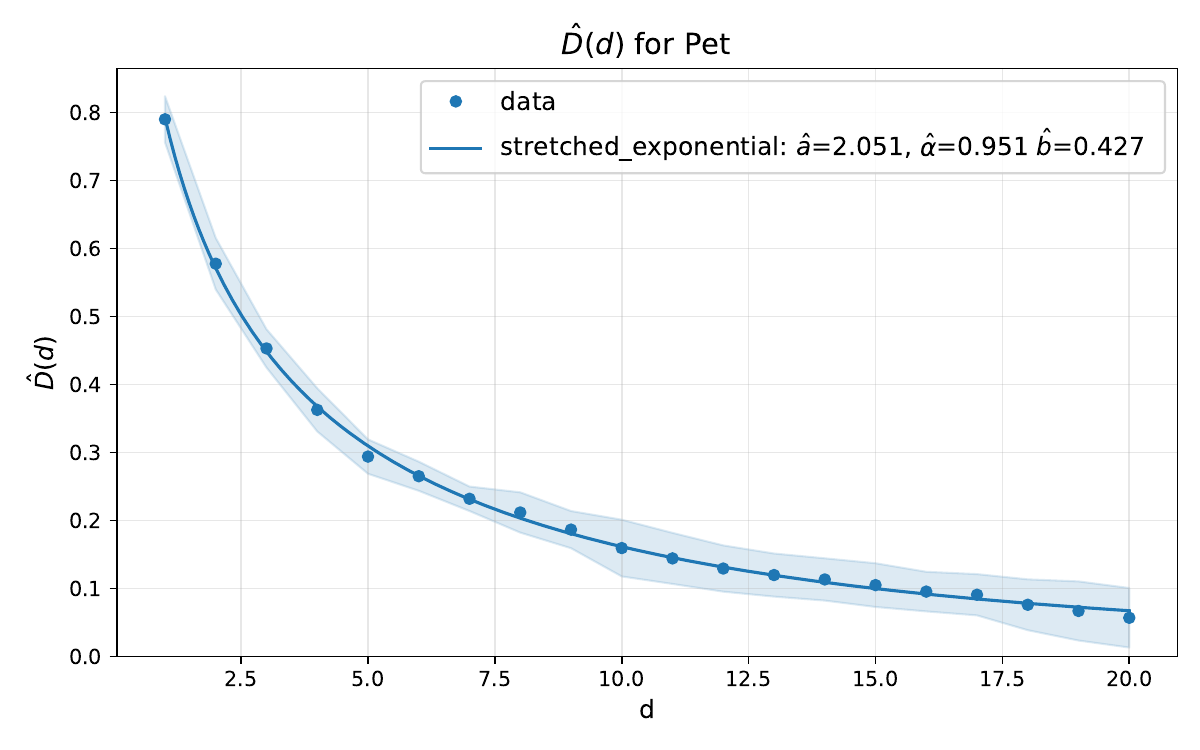}
    \caption{Decay fit $\hat{D}(d)$ for different datasets.}
    \label{fig:decays}
\end{figure}

We estimate the conditional probabilities appearing in~\eqref{eq:Dsigma_bound} by fitting an autoregressive model that takes as input $h(X)$ and a partition $\Omega_b$ for a fixed bin $b$,
and predicts the distribution of the labels ${\sigma = \{Y^{(i)}_{x_b}\}_{x_b \in \Omega_b}}$.
To this aim, we model the joint distribution of the binary segmentation mask restricted to the partition ${\Omega_b}$ and conditioned to $h(X)$ via the autoregressive factorisation
\begin{equation}\label{eq:chain}
  \mathbb{P}_\theta\!\left(\sigma \mid h(X)\right)
  =
  \prod_{x_b\in\Omega_b}
    \mathbb{P}_\theta\!\left(\sigma_{x_b} \mid \sigma_{<x_b},h(X)_{x_b}\right).
\end{equation}
Each conditional $\mathbb{P}_\theta(\sigma_{x_b} = 1 \mid \sigma_{<x_b}, h(X)_{x_b})$ is the sigmoid output of a lightweight network trained by maximising the log-likelihood in \eqref{eq:chain} for a set of bins $b$, partitions $\Omega_b$ and $h(X)$.
Given the fitted model, the entries of $\bar D$, defined in \eqref{eq:Dsigma_bound}, are evaluated for each pair $(x_b, y_b) \in \Omega_b\times\Omega_b$ with ${\Gamma(y_b) > \Gamma(x_b)}$ by computing
\begin{equation}\label{eq:Dsigma_hat}
  \hat{D}^{\,\sigma}_{x_b,y_b}
  =
  \left|
    \mathbb{P}_\theta\!\left(\sigma_{y_b}=1\mid \sigma_{<x_b},\,\sigma_{x_b}=1,\,h(X)_{y_b}\right)
    -\mathbb{P}_\theta\!\left(\sigma_{y_b}=1\mid \sigma_{<x_b},\,\sigma_{x_b}=0,\,h(X)_{y_b}\right)
  \right|.
\end{equation}
Because the conditional $\mathbb{P}_\theta(\sigma_y \mid \sigma_{\le x},h(X)_x)$ is a marginal over the intermediate sites $\sigma_{x+1},\dots,\sigma_{y-1}$, we estimate it using Monte Carlo iterations.
Finally, we estimate $\hat{D}_{x,y}$ by sampling a set of random pasts $\sigma_{<x}$ and taking the configuration correspondent to the largest $\hat{D}_{x,y}^\sigma$, obtaining $\hat{D}_{x,y}=\max_{\sigma_{<x}}\hat{D}^{\,\sigma}_{x,y}.$
We compute $\hat{D}_{x,y}$ over a range of positions $x$ and $y$ and empirically observe in \changes{Fig.}~\ref{fig:decays} that 
its entries depend only on the Euclidean distance $d = \|y - x\|_2.$
This indicates that $\hat{D}_{x,y}$ is translation-invariant and depends solely on the relative displacement between positions. 
Therefore, we model the distance dependence using a stretched exponential decay function,
\begin{equation}\label{eq:fit}
  \hat{D}(d)
  =
  \hat{a}\, \exp(-\hat\alpha\, d^{\hat b}),
  \qquad \hat{a},\,\hat\alpha > 0,\hat b>0
\end{equation}
whose fitted parameters provide a closed-form estimate of the full
coupling matrix and can be used to compute the coefficients $S$ and $c$
that appear in Theorem~\ref{thm:improved}.
Fig.~\ref{fig:decays} shows the estimated $\hat{D}(d)$ and its exponential fit for a selection of the datasets used in the experiments.
See Appendix \ref{app:appendix_bound_estimation} for further details.


\section{Experiments}
\label{experiments}
\begin{figure}
    \centering
    \includegraphics[width=0.9\linewidth]{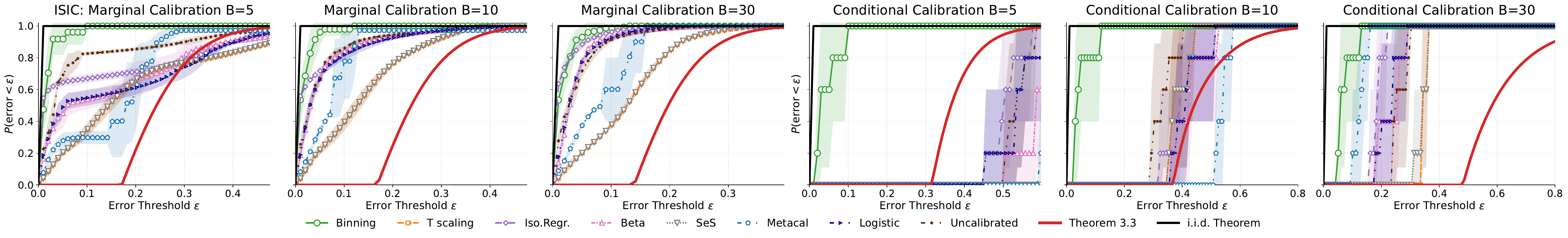}
    \includegraphics[width=0.9\linewidth]{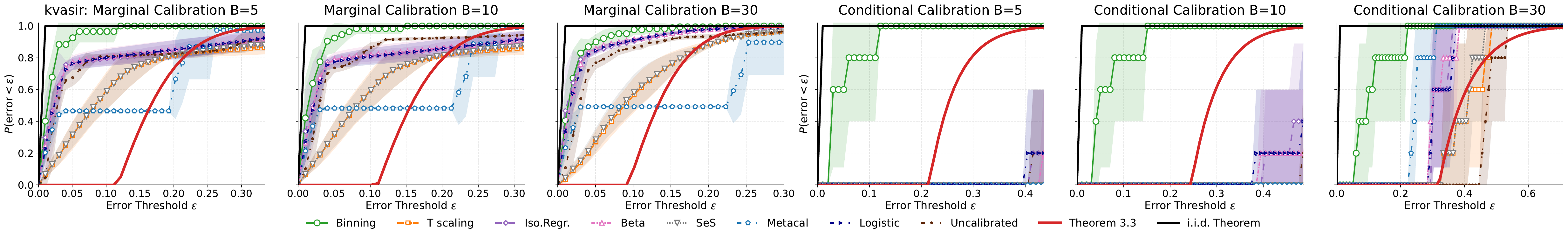}
    \includegraphics[width=0.9\linewidth]{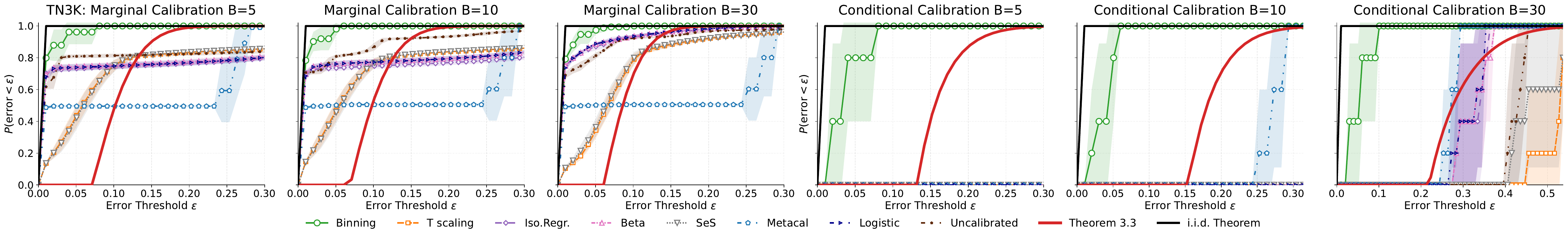}
    \includegraphics[width=0.9\linewidth]{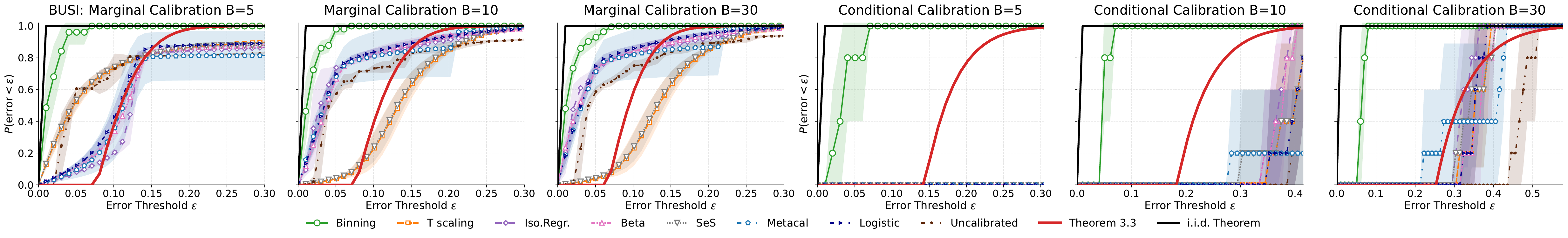}
    \includegraphics[width=0.9\linewidth]{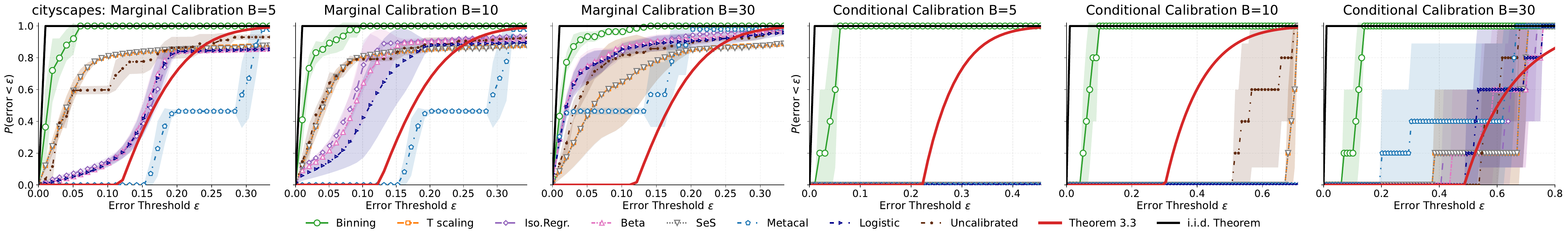}
    \includegraphics[width=0.9\linewidth]{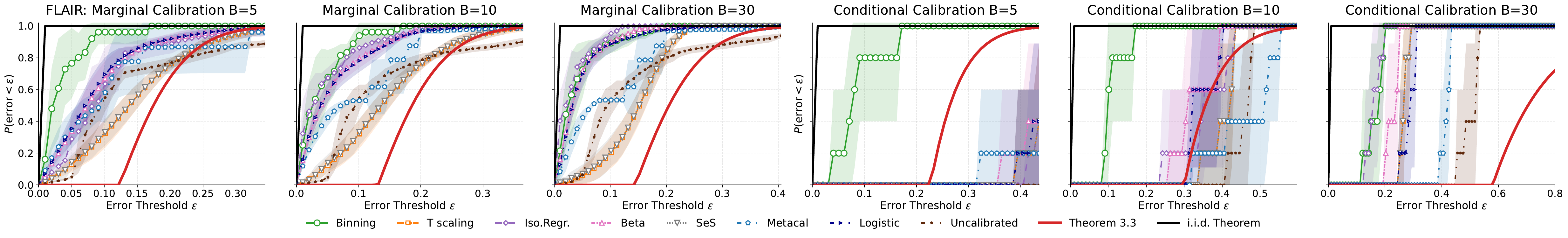}
    \includegraphics[width=0.9\linewidth]{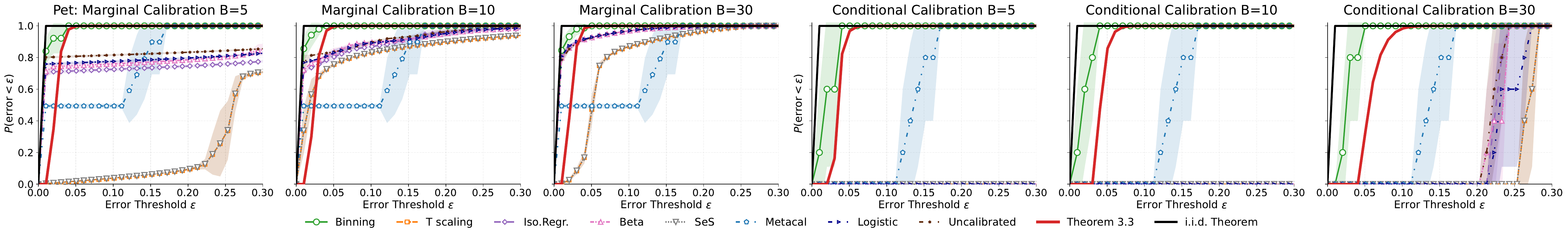}
    \includegraphics[width=0.9\linewidth]{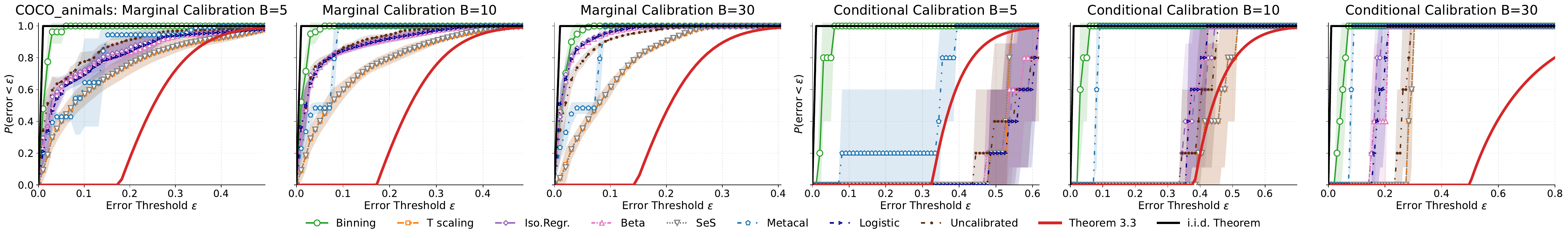}
    \includegraphics[width=0.9\linewidth]{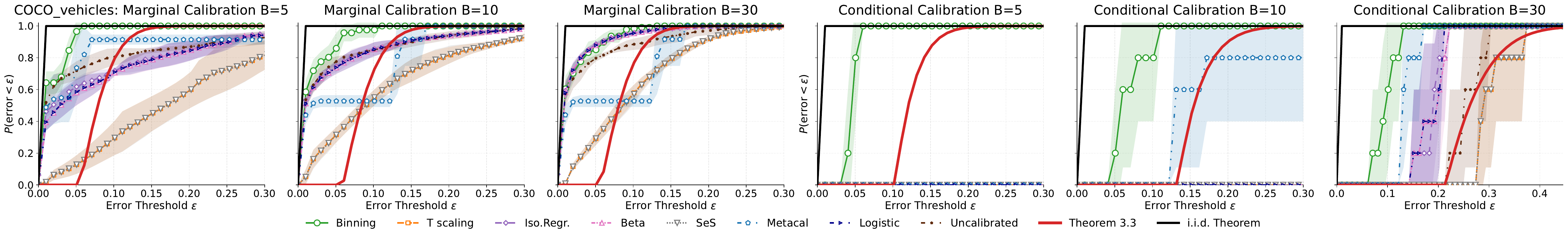}
    \caption{Comparisons of validity curves ${V}_m(\varepsilon), {V}_c(\varepsilon)$ (in red) and empirical ones $\hat{V}_m(\varepsilon), \hat{V}_c(\varepsilon)$ across methods and datasets.}
    \label{fig:results}
\end{figure}

We now show that our theoretically derived bound is consistent with the empirical behavior and compare the binning approach with alternative \textit{post-hoc} calibration methods. 
To do so we compare the true validity curves $V(\varepsilon)$ and the empirical ones $\hat V(\varepsilon)$ both for marginal and conditional calibration. 
Marginal and conditional validity curves are defined as \citep{gupta2021distribution}: 
\[V_m(\varepsilon)=\mathbb{P}(|\E[Y_{{x}}|h(X)_x]-h(X)_{{x}}|\leq\varepsilon),\]
\[ V_c(\varepsilon)=\mathbb{P}(\forall r\in Range(h):|\E[Y_{{x}}|h(X)_{{x}}=r]-r|\leq\varepsilon),\] and are bounded by Theorem \ref{marandcon_cal}. 
Let $\pi^{obs}_b$ the ground-truth observed frequency, 
\begin{equation}\pi^{obs}_b=\frac{\sum^{N_{test}}_{i=1}\sum_{x\in\Omega}Y^{(i)}_{x}\mathbf{1}_{\changes{\{h(X^{(i)})_x\in I_b\}}}}{\sum^{N_{test}}_{i=1}\sum_{x\in\Omega}\mathbf{1}_{\changes{\{h(X^{(i)})_x\in I_b\}}}}.\end{equation}
Then the empirical validity curves can be approximated as
\[\hat V_m(\varepsilon)={\frac{1}{N_{test}\abs{\Omega}}\sum^{N_{test}}_{i=1}\sum^B_{b=1}\sum_{x_b\in\Omega^{(i)}_b}\mathbf{1}_{\{|\pi^{obs}_{b}-h(X^{(i)})_{x_b}|\leq\varepsilon\}}},\]
\[ \hat V_c(\varepsilon)=\mathbf{1}{\Big\{\max_{i=1,\dots,N_{test}}\max_{b=1,\dots,B}\max_{x_b\in\Omega^{(i)}_b}|\pi^{obs}_{b}-h(X^{(i)})_{x_b}|\leq\varepsilon\Big\}}.\]
Both marginal and conditional calibration are estimated as the mean across $5$ random calibration and test splits. The error bars in Figure \ref{fig:results} refer to $\pm1$ standard deviation across those random splits.
\paragraph{Baselines and Datasets} We compare our method with six \textit{post-hoc} calibration baselines. 
\textit{Temperature Scaling} \citep{guo2017calibration} rescales the network logits by a single learned scalar $T>0$. 
\textit{Logistic Scaling} \citep{platt1999probabilistic} fits a logistic function on top of the network outputs, learning an affine transformation of the logits to map them to calibrated probabilities.
\textit{Meta-Cal} \citep{ma2021meta} first learns a ranking function to categorize calibration samples in two groups, then applies temperature scaling to one group and sets the other as random predictions.
\textit{Selective Scaling} \cite{wang2023calibrating} is an extension of Meta-cal where a binary classifier categorizes pixels in two groups scaled with two different temperatures $T_1$ and $T_2$.
\textit{Beta Calibration} \citep{kull2017beta} fits a parametric map based on the Beta distribution family, learning three parameters that control the shape and shift of the probability correction. 
\textit{Isotonic Regression} \citep{zadrozny2002transforming} is a non-parametric method that fits a monotone piecewise-constant mapping via the pool-adjacent-violators algorithm.
In Appendix \ref{app:baselines_methods}, we provide further details on these baseline approaches, along with a discussion of the effect of \textit{post-hoc} calibration on segmentation accuracy (e.g., DICE score).

To assess the generality of our approach, we evaluate all methods across nine segmentation datasets spanning diverse imaging domains and anatomical structures: 
ISIC \citep{gutman2016skin}, Kvasir \citep{Pogorelov:2017:KMI:3083187.3083212}, TN3K \citep{GONG2023106389}, BUSI \citep{tasnim2024cam, al2020dataset}, Cityscapes \citep{Cordts2015Cvprw,Cordts2016Cityscapes}, FLAIR1 \citep{ign2022flair1}, Oxford-IIIT Pet \citep{parkhi12a}, and two subsets of the COCO dataset \citep{lin2014microsoft}. 
Where datasets provided multi-class annotations, predictions were binarized by collapsing all foreground classes into a single positive class. 
The diversity of modalities, domains, and backbone architectures — spanning both medical and non-medical imaging and detailed in Appendix \ref{app:datasets_and_models} — tests our method across highly heterogeneous conditions, providing a broad basis for evaluating its generality and robustness.
\paragraph{Results}
We now compare, and show in Figure \ref{fig:results}, the empirical validity curves $\hat{V}(\varepsilon)$ of the binning strategy, and of the other baselines, with our theoretical bound, as well as with the $i.i.d.$ bound that would apply under the (hypothetical) assumption that all pixels are independent. 
Our experiments highlight two main findings. 
(1) The empirical validity curves of the binning method, evaluated across several datasets, are consistent with our theoretical bound. 
As expected, marginal calibration is easier to satisfy: the binning curve shows good calibration properties, i.e. it attains small calibration errors even for small $\varepsilon$, and the bound consistently controls the binning curve across the experiments.
In some cases, the empirical curve approaches the $i.i.d.$ curve, which can be attributed to an intrinsically good calibration of the pre-trained model. 
Conditional calibration, by contrast, is generally more difficult to achieve. 
Nevertheless, our derivation yields a meaningful and nontrivial bound that is closely tracked by the binning validity curves.
The $i.i.d$ curve further confirms that the independence assumption is too strong for calibrating segmentation models, and would yield incorrect guarantees.
While the theoretical bound is only available in estimated form, these results reinforce the validity of the binning method in this setting and justify our strategy of considering each pixel as a data point.
(2) The binning strategy, supported by our theoretical calibration error bound, outperforms the other \textit{post-hoc} calibration baselines. 
In the marginal setting, the binning approach achieves superior performance relative to most of the competing approaches. 
Isotonic regression, beta calibration and logistic scaling exhibit comparable behavior on most datasets.
Temperature scaling, selective scaling and meta-cal, by contrast, yields unsatisfactory calibration performance on several datasets. 
In the conditional setting, the advantage of our method becomes even more pronounced. 
Specifically, the binning approach is the only method capable of attaining low calibration error (<0.1) with high probability (>90\%).
\changes{Among the baselines in the conditional setting, meta-cal performs best, achieving competitive results on a few datasets, such as Pet and COCO\_animals.}
Furthermore, we observe that increasing the number of bins $B$ tends to slightly reduce the performance of the approach. 
This can be attributed to the fact that, as $B$ grows, fewer points fall within each individual bin, leading to less reliable statistics per bin and thus noisier calibration estimates. 
However, using a larger number of bins also confers the key advantage of more localized and sharper calibration. 
Ultimately, the optimal number of bins is problem-dependent, and the right balance between statistical reliability and calibration sharpness will vary across applications and datasets.
\changes{We provide additional experiments in Appendix \ref{app:addexp}.}

\section{Conclusions and Limitations}
\label{conclusions}
Motivated by the need for calibration guarantees in the context of segmentation, we investigated this problem and obtained the following results. 
(1) We derived an improved theoretical bound for segmentation-like data using concentration inequalities. 
Specifically, we model the segmentation task as a Bernoulli random field, explicitly incorporating the variance into the bound, which leads to a tighter guarantee. 
(2) We revisited the binning approach for binary segmentation calibration, constructing a calibrator that satisfies both marginal and conditional calibration properties, in accordance with the aforementioned theoretical bound. 
Notably, we formally justified treating each pixel as an individual data point even in the absence of $i.i.d.$ assumptions, showing that this choice not only admits rigorous guarantees but also proves effective in practice.
In fact, the binning approach consistently achieves superior performance compared to the baselines, particularly in the conditional setting, where most alternative baselines fail to reach comparable levels of calibration across the majority of datasets.

The primary limitations of the present work are as follows. 
First, our analysis is restricted to binary segmentation; an important direction for future research is the extension of the theoretical framework to multi-class segmentation. 
Second, while our theoretical justification for a binning approach in the segmentation setting remains valid and supported by the experiments,
the bound requires empirical estimation, meaning that the guarantee is evaluated approximately rather than exactly.
\changes{Moreover, deriving the final bound requires taking the supremum over configurations for both $D^\sigma$ and the variance $p_i(1-p_i)$, which is potentially loose.
Although our experiments show that, in general, the resulting bounds remain useful and tight, developing more refined and efficient strategies to sharpen them is a promising direction for future work.}
Finally, we are providing pixel-wise guarantees; a natural question is whether it is possible to bound the calibration error conditioned on specific regions of the image. 

\section*{Acknowledgement}
This research was funded in whole or in part by the Austrian Science Fund (FWF)
10.55776/PIN2770025 and by the BioTechMed-Graz Young
Research Group Grant CICLOPS.

\bibliography{main}
\bibliographystyle{tmlr}

\appendix

\section{Proofs}
\label{app:proofs}

\subsection{Derivation of the martingale increment identity}
\label{A:Vi}
For a Bernoulli alphabet $A=\{0,1\}$, the martingale increment, using the law of total expectation, reduces to:
\begin{align*}
  V_i(\sigma) &= \E[g \mid \sigma_{<i}, \sigma_i] - {\bigl[p_i \E[g \mid \sigma_{<i},\, \sigma_i = 1] + (1-p_i)\E[g \mid \sigma_{<i},\, \sigma_i = 0]\bigr]}\\
      &= \begin{cases}
           (1-p_i)\Delta_i & \text{if } \sigma_i = 1, \\
           -p_i\,\Delta_i   & \text{if } \sigma_i = 0.
         \end{cases}
\end{align*}
Both cases are captured by the identity $V_i(\sigma) = (\sigma_i - p_i)\,\Delta_i$ of \eqref{eq:Vi}.

\subsection{Proof of Lemma \ref{lem:bennett}}
\label{pr:lemmabenent}
\begin{proof}
We first notice that if $\Delta_i = 0$ both sides equal zero and the bound holds trivially. In the general case,
fix $\F_{i-1}$ and recall $p_i$, $\Delta_i$ are
$\F_{i-1}$-measurable.
Given $\F_{i-1}$, $W_i := \frac{V_i}{\Delta_i}= \sigma_i - p_i$ is a centered two-point
random variable with:
\[
  \E[W_i\mid\F_{i-1}] = 0,\quad
  W_i \leq 1-p_i \leq 1 =: b,\quad
  \E[W_i^2\mid\F_{i-1}] = p_i(1-p_i) =: v.
\]
Apply Theorem~2.9 of \citep{Massart} with $n=1$, $b=1$,
$v = p_i(1-p_i)$, and parameter $\lambda\Delta_i$ in place of $\lambda$:
\[
  \log\E\bigl[e^{\lambda\Delta_i W_i}\mid\F_{i-1}\bigr]
  \;\leq\; \frac{v}{b^2}\,\phi(b\cdot\lambda\Delta_i)
  \;=\; p_i(1-p_i)\,\phi(\lambda\Delta_i).
\]
Since $V_i = \Delta_i W_i$, this is exactly \eqref{eq:bennett-cond}.
We use $b=1$ rather than the tighter $b=1-p_i$ to keep the bound symmetric in $p_i$.
\end{proof}

\subsection{Derivation of the $\Delta_i$ bound}
\label{A:delta}
By \citet[Eq.~5]{chazottes2007}, $\Delta_i(\sigma)\leq(D^\sigma\delta g)_i$, where the matrix-vector product is
\[
(D^\sigma\delta g)_i := \sum^{n}_{\changes{j\geq i}}D^\sigma_{i,j}\,\delta_j g.
\]
Since $\delta_j g = 1/n$ for all $j$ \eqref{eq:gdef}, this simplifies to
\[
(D^\sigma\,\delta g)_i = \frac{1}{n}\sum_{\changes{j\geq i}}^{n} D^\sigma_{i,j}
=: \frac{R^\sigma_i}{n},
\]
where $R^\sigma_i$ denotes the $i$-th row sum of $D^\sigma$. Taking the supremum over all configurations and setting ${\bar{D}_{i,j} := \sup_\sigma D^\sigma_{i,j}}$ yields
\[
\Delta_i \leq  (D^\sigma\,\delta g)_i = \frac{1}{n}\sum_{\changes{j\geq i}}^{n} {D}^\sigma_{i,j} \leq \sup_\sigma  \frac{1}{n}\sum_{\changes{j\geq i}}^{n} {D}^\sigma_{i,j}=\frac{1}{n}\sum_{\changes{j\geq i}}^{n} \bar{D}_{i,j}=\frac{R_i}{n}.
\]

\subsection{Proof of Theorem \ref{thm:improved}}
\label{pr:proofofthmimproved}
\begin{proof}
From \eqref{eq:martingaledecomp} and \eqref{eq:Vi} we have that:
\[
  g - \E g = \sum_{i=1}^n V_i, \qquad
  V_i = (\sigma_i - p_i)\,\Delta_i,
\]
where both $p_i$, $\Delta_i$ are
$\F_{i-1}$-measurable.
We apply the exponential Chebyshev inequality, and obtain that for any $\lambda > 0$:
\begin{equation}\label{eq:cheby}
  \mathbb{P}(g - \E g \geq \varepsilon)
  \;\leq\; e^{-\lambda \varepsilon}\,\E\!\left[e^{\lambda\sum_{i=1}^n V_i}\right].
\end{equation}
Focusing on the expectation, we peel off one term using the tower property:
\[
  \E\!\left[e^{\lambda\sum_i V_i}\right]
  = \E\!\left[\E\!\left[e^{\lambda V_n}\mid\F_{n-1}\right]
    e^{\lambda\sum_{i=1}^{n-1}V_i}\right].
\]
Now we apply Lemma~\ref{lem:bennett} to bound $\E[e^{\lambda V_n}\mid\F_{n-1}]$.
Notice that $g$ defined in \eqref{eq:gdef} is increasing in $\sigma$, therefore $\Delta_n>0$ and Lemma~\ref{lem:bennett} can be applied.
\changes{More generally, for the conditional MGF of any increment $V_i$, Lemma~\ref{lem:bennett} gives $\E[e^{\lambda V_i}\mid\F_{i-1}]\le\exp\!\big(p_i(1-p_i)\,\phi(\lambda\Delta_i)\big)$. 
The bounding factor $\exp\!\big(p_i(1-p_i)\,\phi(\lambda\Delta_i)\big)$ is itself a random variable, since $p_i$ and $\Delta_i$ are $\F_{i-1}$-measurable; 
consequently it cannot be factored out of the successive conditioning. 
We remove this randomness in two steps, {before} peeling off the terms.}

\changes{First, we bound the increment. For $g = n^{-1}\sum^n_{i=1}\sigma_i$ we have $\delta_j g = 1/n$ for all $j$, so \eqref{eq:deltabound} yields the deterministic bound $\Delta_i \leq (\bar D\,\delta g)_i = R_i/n$. Since $\phi$ is increasing on $[0,\infty)$ and $p_i(1-p_i)\ge 0$, conditionally on $\F_{i-1}$,
\[
  \E\!\left[e^{\lambda V_i}\mid\F_{i-1}\right]
  \;\le\; \exp\!\big(p_i(1-p_i)\,\phi(\lambda\Delta_i)\big)
  \;\le\; \exp\!\Big(p_i(1-p_i)\,\phi\big(\tfrac{\lambda R_i}{n}\big)\Big),
\]
so that the argument of $\phi$ is now the deterministic quantity $\lambda R_i/n$.}

\changes{Second, we remove the conditional variance $p_i(1-p_i)$ using the same strategy applied to the coupling matrix $D^\sigma$. Since $p_i(1-p_i)\le\bar q_i$ by definition of $\bar q_i$ and $\phi\ge 0$,
\[
  \E\!\left[e^{\lambda V_i}\mid\F_{i-1}\right]
  \;\le\; \exp\!\Big(\bar q_i\,\phi\big(\tfrac{\lambda R_i}{n}\big)\Big).
\]
The right-hand side is now a deterministic constant, so it factors cleanly out of each conditional expectation. 
Peeling the $n$ terms one at a time yields}
\begin{equation}\label{eq:succond}
  \log\E\!\left[e^{\lambda(g-\E g)}\right]=
  \log\E\!\left[e^{\lambda\sum_i V_i}\right]
  \;\leq\; \changes{\sum_{i=1}^n \bar q_i\,\phi\!\left(\tfrac{\lambda R_i}{n}\right).}
\end{equation}
It is possible to show, using Taylor expansions, that for $0 \leq u < 3$, the inequality
\begin{equation}\label{eq:phi-bound}
  \phi(u) \;\leq\; \frac{u^2/2}{1 - u/3}
\end{equation}
holds. Restricting to $\lambda < 3n/R_{\max}$ we ensure
$\lambda R_i/n \leq \lambda R_{\max}/n < 3$ for all $i$. We apply
\eqref{eq:phi-bound} with $u = \lambda R_i/n$:
\[
  \changes{\sum_{i=1}^n \bar q_i\,\phi\!\left(\frac{\lambda R_i}{n}\right)}
  \;\leq\;
  \frac{\lambda^2}{2}\sum_{i=1}^n
    \frac{\changes{\bar q_i}\,R_i^2/n^2}{1-\lambda R_i/(3n)}.
\]
Now, we bound the denominator uniformly by its worst case, $1 - \lambda R_{\max}/(3n)$, and obtain:
\begin{equation}\label{eq:subgamma}
  \log\E\!\left[e^{\lambda(g-\E g)}\right]
  \;\leq\;
  \frac{\lambda^2 S/2}{1 - \lambda c}
  \;=:\; \psi(\lambda),
  \qquad for \quad 0 < \lambda < 1/c,
\end{equation}
with $S$ and $c$ as defined in \eqref{eq:BC}.
(This is the sub-gamma form $\psi(\lambda) = \lambda^2 S/(2(1-c\lambda))$
of \citet[Section~2.4]{Massart}.)
We plug in \eqref{eq:subgamma} in \eqref{eq:cheby} and obtain:
\begin{equation}\label{eq:wheretoputlam}\mathbb{P}(g - \E g \geq \varepsilon)
  \;\leq\; e^{-\lambda \varepsilon}e^{\frac{\lambda^2 S/2}{1 - \lambda c}},\qquad for \quad 0 < \lambda < 1/c. \end{equation}
In order to make the right-hand side minimal, we derive by $\lambda$ and obtain:
\[\frac{d}{d\lambda}\Big(-\lambda \varepsilon +\frac{\lambda^2 S/2}{1 - \lambda c}\Big)=-\varepsilon+\frac{\lambda S(2-\lambda c)}{2(1-\lambda c)^2}\]
Setting the derivative to zero gives a quadratic equation in $\lambda$, selecting the $\lambda$ that satisfies $
\lambda<\frac{1}{c}$ gives us:
\[\lambda^* = \frac{1}{c}\Big(1-\sqrt{\frac{S}{S+2\varepsilon c}}\Big)\]
By substituting $\lambda^*$ in \eqref{eq:wheretoputlam} we get the following bound:
\[
  \log \mathbb{P}(g - \E g \geq \varepsilon)
  \;\leq\; -\frac{S}{c^2}\eta\Big(\frac{c\varepsilon}{S}\Big),\quad \eta(x)=1+x-\sqrt{1+2x}
\]

The two-sided bound \eqref{eq:main} follows selecting $-g$. 
In this case, $-g$ is decreasing in $\sigma$, therefore $\Delta_i$ has to be defined as $\Delta_i=\E[g \mid \sigma_{<i},\, \sigma_i = 0] - \E[g \mid \sigma_{<i},\, \sigma_i = 1]$. 
Notice that using $-g$ swaps $p_i$  and $1-p_i$ but
leaves $S$ and $c$ unchanged since $p_i(1-p_i)$ is symmetric.
\end{proof}

\subsection{Extension of Theorem \ref{thm:improved} to $\Z^2$}
\label{subsec:extensionZ2}
To extend Theorem \ref{thm:improved} to $\Z^2$ we follow the same strategy of \citep[Section~2.4]{chazottes2007}.
Let $\Omega\subset \Z^2$ and $\Lambda=\{0,1\}^\Omega$. 
Introduce any enumeration $\Gamma:\Z^2\to\N$. We use the notation ${(\leq x)=\{y\in\Z^2:\Gamma(y)\leq\Gamma(x)\}}$, which is the $\Z^2$ equivalent of ($\leq i$) in the $\N$ case of Theorem \ref{thm:improved}.
Therefore, $\mathcal{F}_{\leq x}$ denotes the sigma-field generated by $\sigma_y,y\leq x$ and $\mathcal{F}_0$ is the trivial sigma-field.
Analogously to \eqref{eq:martingaledecomp} we have the following decomposition:
\begin{equation}
  g(\sigma)-\E[g]=\sum_{x\in\Omega}V_x(\sigma),\qquad \text{where}\quad V_x(\sigma):=\E[g|\mathcal{F}_{\leq x}]-\E[g|\mathcal{F}_{<x}].
\end{equation}
Finally we introduce the coupling matrices as
\begin{equation}
  D^{\sigma}_{x,y}:=\mathbb{P}^{\sigma_{<x}}_{x,1,0}\Big(\sigma^{(1)}_y\neq\sigma^{(2)}_y\Big),\quad \bar{D}_{x,y}=\sup_\sigma D^\sigma_{x,y},
  \label{eq:DinZ2}
\end{equation}
where $\mathbb{P}^\sigma_{x,1,0}$ denotes the maximal coupling between the conditional measures $\mathbb{P}(\cdot|\sigma_{<x},\sigma_x=1)$ and $\mathbb{P}(\cdot|\sigma_{<x},\sigma_x=0)$.
\begin{theorem}[Improved concentration for Bernoulli mean in $\Z^2$]%
\label{thm:improvedZ2}
Let $\Omega\subset \Z^2$ and let ${g = |\Omega|^{-1}\sum_{x\in\Omega}\sigma_x}$.
Define $\bar D_{x,y}$ as in \eqref{eq:DinZ2}, ${R_x := \sum_{\changes{y\geq x}} \bar D_{x,y}}$, and 
\changes{$\bar q_x := \sup_{\sigma_{<x}}p_x(1-p_x)$ with}
${p_x := \mathbb{P}(\sigma_x=1\mid\sigma_{<x})}$.
Set
\begin{equation}\label{eq:BCZ2}
  S \;:=\; \frac{1}{|\Omega|^2}\sum_{x\in\Omega} \changes{\bar q_x}\,R_x^2,
  \qquad
  R_{\max} := \max_{x\in\Omega} R_x, \quad c \;:=\; \frac{R_{\max}}{3|\Omega|}.
\end{equation}
Then for all $\varepsilon > 0$:
\begin{equation}\label{eq:mainZ2}
  \boxed{\mathbb{P}(\abs{g - \E[g]} \geq \varepsilon)
  \;\leq\;
  2\exp\!\left(-\frac{S}{c^2}\eta\Big(\frac{c\varepsilon}{S}\Big)\right)}\qquad with \quad \eta(x)=1+x-\sqrt{1+2x}.
\end{equation}
\end{theorem}
\begin{proof}
 The enumeration $\Gamma$ maps the indices of $\Z^2$ to $\N$. Therefore, we can apply the proof of Theorem \ref{thm:improved} and obtain the bound \eqref{eq:mainZ2}.
\end{proof}

\subsection{Proofs of Calibration Theorems}
\label{A:cal}

\begin{proof}[Proof of Theorem~\ref{ourbound1}]
Fix a bin $b\in\{1,\dots,B\}$. The observed frequency can be written as
\[
  \hat\pi_b = \frac{1}{N_b}\sum^{N_{cal}}_{i=1}\sum_{x_b\in\Omega^{(i)}_b} Y^{(i)}_{x_b},
\]
which has the form $g = N_b^{-1}\sum_i \sigma_i$, i.e.\ the empirical mean of $N_b$ Bernoulli variables indexed over the disjoint domains $\Omega^{(1)}_b,\dots,\Omega^{(N_{cal})}_b$.
Moreover, since the binning event $E_{\mathcal{B}}$ is fixed, $N_b$ is deterministic and the randomness is over the Bernoulli variables.

Since images are drawn independently, pixels from different images are independent. 
Therefore the coupling matrix $\bar D$ of the full collection of $N_b$ pixels is block-diagonal, where each block $\bar{D}^{(i)}$ correspond to the coupling matrix of image $i$ conditioned on the binning event $E_{\mathcal{B}}$ (i.e. the collection $\{Y^{(i)}_{x_b}\}_{x_b\in\Omega^{(i)}_b}$). 
$\bar D_{x_b,y_b}$ is defined as in Section~\ref{subsec:Dg} for pixels $x_b,y_b$ within the same image $i$, and $\bar D_{x_b,y_b}=0$ for pixels from different images.
As a consequence, the row sums of the global coupling matrix reduce to the within-image row sums:
\[
  R^{(i)}_{x_b} = \sum_{\changes{y_b\geq x_b}} \bar D^{(i)}_{x_b,y_b},
\]
since all cross-image terms vanish. The quantities $S_b$ and $c_b$ are computed by substituting into \eqref{eq:BCZ2}:
\[
  \changes{\bar{q}_{x_b}^{(i)}=\sup_{Y^{(i)}_{<{x_b}}}p^{(i)}_{x_b}(1-p^{(i)}_{x_b})},\quad
  S_b = \frac{1}{N_b^2}\sum^{N_{cal}}_{i=1}\sum_{x_b\in\Omega^{(i)}_b} \changes{\bar{q}_{x_b}^{(i)}}\,(R^{(i)}_{x_b})^2, \qquad
  c_b = \frac{R_{\max,b}}{3N_b},
\]
with $R_{\max,b} = \max^{N_{cal}}_{i=1}\max_{x\in\Omega^{(i)}_b} R^{(i)}_{x_b}$.

We can now apply Theorem~\ref{thm:improvedZ2}, conditioned on the binning event $E_{\mathcal{B}}$ to fix $N_b$, with $\sigma=\{Y_{x_b}^{(i)}\}^{N_{cal}}_{i=1,  x_b\in\Omega^{(i)}_b}$, $|\Omega|=N_b$, $g=\hat\pi_b$:
\[
  \mathbb{P}(\abs{\hat\pi_b - \E[\hat{\pi}_b\mid E_{\mathcal{B}}]} \geq \varepsilon \mid E_{\mathcal{B}})
  \leq 2\exp\!\left(-\frac{S_b}{c_b^2}\,\eta\!\left(\frac{c_b\,\varepsilon}{S_b}\right)\right).
\]
Marginalising over all the binning events we get
\[
  \mathbb{P}(\abs{\hat\pi_b - \E[\hat{\pi}_b\mid E_{\mathcal{B}}]} \geq \varepsilon)
  \leq 2\exp\!\left(-\frac{S_b}{c_b^2}\,\eta\!\left(\frac{c_b\,\varepsilon}{S_b}\right)\right). \qedhere
\]
\end{proof}

\begin{proof}[Proof of Theorem~\ref{ourbound2}]
By a union bound over the $B$ bins:
\begin{equation*}\begin{aligned}
  \mathbb{P}(\exists\, b\in[B]: \abs{\hat\pi_b - \E[\hat{\pi}_b\mid E_{\mathcal{B}}]} \geq \varepsilon)
  &\leq \sum^B_{b=1} \mathbb{P}(\abs{\hat\pi_b - \E[\hat{\pi}_b\mid E_{\mathcal{B}}]} \geq \varepsilon)\\
  &\leq \sum^B_{b=1} 2\exp\!\left(-\frac{S_b}{c_b^2}\,\eta\!\left(\frac{c_b\,\varepsilon}{S_b}\right)\right).
\end{aligned}\end{equation*}
\end{proof}

\begin{proof}[Proof of Theorem~\ref{marandcon_cal}]
\textit{Marginal calibration.}
We start by noticing that conditioning on the event, $E_{\mathcal{B}}:=\{f(X^{(1)})_x\in I_b, f(X^{(1)})_z\in I_b,\dots,f(X^{(N_{cal})})_x\in I_b,\dots\}$, is not equivalent to conditioning on the event $E_h:=\{h(X^{(1)})_x,\dots,h(X^{(N_{cal})})_x,\dots\}.$
The two events coincide only if $\hat{\pi}_{b'}\neq \hat{\pi}_{b}\ \forall b'\neq b$.
In that case knowing that pixel $x$ has an associated calibrated predictor $h(X)_x$, and therefore $\hat\pi_b$, automatically tells that $f(X)_x\in I_b$, since $\hat\pi_b$ is different for each bin $b$.
As introduced in \citet{gupta2021distribution}, a simple trick to not lose the informativeness of $h$ is to randomise it with an arbitrary small value.
More formally, define the new calibrated predictor $\tilde{h}(X)_x$ as
\[\tilde{h}(X)_x=\sum^B_{b=1}\mathbf{1}_{\{f(X)_x\in I_b\}}({\hat\pi_b+\delta U_b}),\quad U_b\sim\mathcal{U}[-1,1],\]
where $\delta$ is an arbitrary parameter. 
If we choose, for example, $\delta=10^{-20}$, then the new randomized calibrated predictor is in practice indistinguishable from the old one, but, we can now write that $\E[\hat\pi_b\mid E_{\mathcal{B}}]=\E[\hat\pi_b\mid E_{\tilde{h}}].$
In the following, we denote as $\tilde{h}(X)$ the union of all of the pixel-wise events $\tilde{h}(X)_x\ \forall x\in\Omega$.

We proceed by computing the conditional expected value of $\hat\pi_b$ as
\begin{equation*}
  \begin{aligned}
    \E[\hat\pi_b\mid E_{\mathcal{B}}]&=\frac{1}{N_b}\E\bigg[\sum^{N_{cal}}_{i=1}\sum_{x\in\Omega}Y^{(i)}_x\mathbf{1}_{{\{f(X^{(i)})_x\in I_b\}}}\mid \tilde{h}(X^{(1)}),\dots,\tilde{h}(X^{(N_{cal})})\bigg]\\
                       &=\frac{1}{N_b}\sum^{N_{cal}}_{i=1}\sum_{x\in\Omega}\mathbf{1}_{{\{f(X^{(i)})_x\in I_b\}}}\E[Y^{(i)}_x\mid \tilde{h}(X^{(1)}),\dots,\tilde{h}(X^{({N_{cal}})})]\\
                       &=\frac{1}{N_b}\sum^{N_{cal}}_{i=1}\sum_{x_b\in\Omega^{(i)}_b}\E[Y^{(i)}_{x_b}\mid \tilde{h}(X^{(i)})]\ \ \ \quad \quad\bigg(\{X^{(i)}\}^{N_{cal}}_{i=1}\ i.i.d.\bigg)\\
                       &=\frac{1}{N_b}\sum^{N_{cal}}_{i=1}\sum_{x_b\in\Omega^{(i)}_b}\E[Y^{(i)}_{x_b}\mid \tilde{h}(X^{(i)})_{x_b}]\ \ \ \quad\bigg(Y^{(i)}_{x_b}\mid \tilde{h}(X^{(i)})_{x_b}\perp\!\!\!\perp \tilde{h}(X^{(i)})_{y_b},\ y_b\neq x_b\bigg)\\
                       &=\frac{1}{N_b}\sum^{N_{cal}}_{i=1}\sum_{x_b\in\Omega^{(i)}_b}\E[Y_{x_b}\mid \tilde{h}(X)_{x_b}]\ \quad\qquad\bigg(\{X^{(i)}\}^{N_{cal}}_{i=1}\ i.i.d.\bigg)\\
                       &=\E[Y_{x_b}\mid \tilde{h}(X)_{x_b}]\frac{1}{N_b}\sum^{N_{cal}}_{i=1}\abs{\Omega^{(i)}_b}\qquad\qquad\bigg(\E[Y_{x_b}\mid \tilde{h}(X)_{x_b}]=\E[Y_{y_b}\mid \tilde{h}(X)_{y_b}]\ \forall x_b,y_b\bigg)\\
                       &=\E[Y_{x_b}\mid \tilde{h}(X)_{x_b}]\qquad\qquad\qquad\qquad\qquad\bigg(\sum^{N_{cal}}_{i=1}\abs{\Omega^{(i)}_b}=N_b\bigg)
  \end{aligned}
\end{equation*} 
Since the derivation holds for each bin $b$, we drop the $x_b$ notation and we continue for a generic pixel $x$.
Substituting $\E[Y_x\mid \tilde{h}(X)_x]$ in Theorem \ref{ourbound1} we obtain:
\[\mathbb{P}\bigg(\abs{\hat\pi_b-\E[Y_x\mid \tilde{h}(X)_x]}\geq\varepsilon\bigg)\leq2\exp\!\left(-\frac{S_b}{c_b^2}\,\eta\!\left(\frac{c_b\, \varepsilon}{S_b}\right)\right).\]
We notice that by definition, for a fixed bin $b$, $\hat\pi_b+\delta U_b=\tilde{h}(X)_x$ and therefore \[{\abs{\tilde{h}(X)_x-\E[Y_x\mid \tilde{h}(X)_x]}=\abs{\hat\pi_b +\delta U_b -\E[Y_x\mid \tilde{h}(X)_x]}}.\] 
Using the triangle inequality and the definition of $U_b$ we get
\[\abs{\tilde{h}(X)_x-\E[Y_x\mid \tilde{h}(X)_x]}\leq\abs{\hat\pi_b-\E[Y_x\mid \tilde{h}(X)_x]}+\delta.\]
We proceed by noticing that
\[\abs{\tilde{h}(X)_x-\E[Y_x\mid \tilde{h}(X)_x]}\geq\varepsilon\implies \abs{\hat\pi_b-\E[Y_x\mid \tilde{h}(X)_x]}\geq\varepsilon-\delta,\]
and therefore
\begin{equation*}
	\begin{aligned}
		\mathbb{P}\bigg(\abs{\tilde{h}(X)_x-\E[Y_x\mid \tilde{h}(X)_x]}\geq\varepsilon\bigg)&\leq\mathbb{P}\bigg(\abs{\hat\pi_b-\E[Y_x\mid \tilde{h}(X)_x]}\geq\varepsilon-\delta\bigg)\\ &\leq2\exp\!\left(-\frac{S_b}{c_b^2}\,\eta\!\left(\frac{c_b\, (\varepsilon-\delta)}{S_b}\right)\right).
	\end{aligned}
\end{equation*}
Taking the complement we obtain the final result
\[
  \mathbb{P}\bigg(\abs{\E[Y_x\mid \tilde{h}(X)_x] - \tilde{h}(X)_x} \leq \varepsilon\bigg) \geq 1 - 2\exp\!\left(-\frac{S_b}{c_b^2}\,\eta\!\left(\frac{c_b(\varepsilon-\delta)}{S_b}\right)\right) = 1 - \alpha_{1,b}.
\]

\textit{Conditional calibration.}
If we take the complement of Theorem~\ref{ourbound2} we obtain
\[\mathbb{P}(\forall\, b\in[B]:\abs{\hat\pi_b-\E[\hat\pi_b\mid E_{\mathcal{B}}]}\leq \varepsilon)\geq1- 2\sum^B_{b=1}\exp\!\left(-\frac{S_b}{c_{{b}}^2}\,\eta\!\left(\frac{c_{b} \varepsilon}{S_{b}}\right)\right).\]
Moreover, by the same argument as in the marginal case
\[\mathbb{P}\bigg(\forall\, b\in[B]:\abs{\tilde{h}(X)_x-\E[Y_x\mid \tilde{h}(X)_x]}\leq \varepsilon\bigg)\geq1- 2\sum^B_{b=1}\exp\!\left(-\frac{S_b}{c_{{b}}^2}\,\eta\!\left(\frac{c_{b} (\varepsilon-\delta)}{S_{b}}\right)\right).\]
After the randomization $\tilde{h}(X)_x=\hat\pi_b+\delta U_b$ introduced above, the values $\hat\pi_1+\delta U_1,\dots,\hat\pi_B+\delta U_B$ are pairwise distinct almost surely, so the range of $\tilde{h}(X)_x$ coincides with $\{\hat\pi_b+\delta U_b\}^B_{b=1}$ and to each $r$ in this range corresponds a unique bin $b(r)\in[B]$ with $\hat\pi_{b(r)}+\delta U_{b(r)}=r$. 
Then
\[\mathbb{P}\bigg(\forall\, r\in Range(h):\abs{\E[Y_x\mid \tilde{h}(X)_x=r]-r}\leq\varepsilon\bigg)\geq 1-2\sum^B_{b=1}\exp\!\left(-\frac{S_b}{c_{{b}}^2}\,\eta\!\left(\frac{c_{b}(\varepsilon-\delta)}{S_{b}}\right)\right)=1-\alpha_2. \qedhere\]
\end{proof}

\section{Baseline Methods}
\label{app:baselines_methods}
We now describe the baseline methods. 
For all the \textit{post-hoc} baselines (and our method), the spatial predictions (e.g. softmax and logits maps
of shape $N \times H \times W$) are flattened before fitting, and each pixel is treated independently from the others.
 
\subsection{Temperature Scaling}
\label{app:temp_scaling}
Temperature Scaling~\citep{guo2017calibration} is the simplest post-hoc
calibration scheme.  Given raw logits $\mathbf{z} \in \mathbb{R}^{C}$ produced
by the network, the calibrated probability vector is
 
\begin{equation}
    \hat{\mathbf{p}} = \operatorname{softmax}\!\left(\frac{\mathbf{z}}{T}\right),
    \label{eq:temp_scaling}
\end{equation}
 
where $T > 0$ is a single scalar temperature shared across all classes and all
samples.  For binary classification, \eqref{eq:temp_scaling} reduces to
 
\begin{equation}
    \hat{p} = \sigma\!\left(\frac{z}{T}\right),
\end{equation}
 
where $\sigma$ is the sigmoid function and $z$ is the scalar logit. $T$ is found by minimizing the cross-entropy
on the calibration set:
 
\begin{equation}
    T^* = \arg\min_{T > 0}\;
          \frac{1}{N}\sum_{i=1}^{N}
          \ell_{\mathrm{CE}}\!\left(
              \operatorname{softmax}\!\left(\mathbf{z}_i / T\right),\, y_i
          \right).
\end{equation}
 
In practice, over-confident networks yield $T^* > 1$, which softens the output distribution toward uniform. 
In the experiments, the temperature is log-parametrized as $T = e^{\tau}$ with $\tau \in \mathbb{R}$,
ensuring positivity throughout optimization without any explicit constraint.
The single parameter $\tau$ is optimized with L-BFGS using a strong Wolfe
line search.
 

\subsection{Logistic scaling}
Logistic scaling~\citep{platt1999probabilistic} (also known as Platt
scaling) learns an affine transformation of the raw logits followed by a
sigmoid, producing calibrated probabilities via the map

\begin{equation}
    \hat{p} = \sigma(a \cdot z + b),
    \label{eq:logistic_map}
\end{equation}

where $z \in \mathbb{R}$ is the pre-sigmoid logit, $a > 0$ is a scale
parameter, and $b \in \mathbb{R}$ is a bias term. 

Logistic scaling generalises Temperature Scaling, which corresponds to the
special case $a = 1/T$ and $b = 0$.  Conversely, it is a special case of Beta
Calibration~\citep{kull2019beyond}, which uses the richer map
$\hat{p} = \sigma(a\,\log p - b\,\log(1-p) + c)$ with independent parameters
for $\log p$ and $\log(1-p)$.

Given calibration pairs $\{(z_i, y_i)\}_{i=1}^{N}$ with $y_i \in \{0,1\}$,
the parameters are found by minimising the binary cross-entropy:

\begin{equation}
    a^*,\, b^* = \arg\min_{a > 0,\, b \in \mathbb{R}}\;
    \frac{1}{N}\sum_{i=1}^{N}
        \ell_{\mathrm{CE}}\!\bigl(\sigma(a\, z_i + b),\, y_i\bigr).
    \label{eq:logistic_fit}
\end{equation}

In the experiments, the positivity constraint on~$a$ is enforced via a log-parameterisation
$a = e^{\alpha}$, $\alpha \in \mathbb{R}$, and the two-parameter problem is
solved with L-BFGS.
 
\subsection{Beta Calibration}
\label{app:beta_calibration}
Beta Calibration~\citep{kull2019beyond} generalizes logistic (Platt) scaling for
binary classifiers by exploiting the Beta family of distributions. The calibration map
is
 
\begin{equation}
    \hat{p} = \sigma\!\bigl(a\log p - b\log(1-p) + c\bigr),
    \label{eq:beta_map}
\end{equation}
 
where $a,b > 0$ are shape parameters and $c \in \mathbb{R}$ is a bias.
Equation \ref{eq:beta_map} can be derived as the ratio of two Beta
densities evaluated at $p$. The three-parameter family
subsumes temperature scaling ($a = b$, $c = 0$) and Platt scaling ($a = b$,
$c$ free) as special cases, providing strictly greater expressiveness.
The parameters are optimized by minimizing binary cross-entropy on the
calibration set:
 
\begin{equation}
    (a^*, b^*, c^*) = \arg\min_{a,b>0,\,c}
          \frac{1}{N}\sum_{i=1}^{N}
          \ell_{\mathrm{BCE}}\!\left(\hat{p}_i,\, y_i\right).
\end{equation}

In the experiments, the positivity constraints on $a$ and $b$ are handled via log-parametrization
($a = e^{\alpha}$, $b = e^{\beta}$), reducing the constrained problem to an
unconstrained three-parameter optimization over $(\alpha, \beta, c)$.
L-BFGS with a strong Wolfe line search is used. 

 
\subsection{Isotonic Regression Calibration}
\label{app:isotonic}
Isotonic Regression Calibration~\citep{zadrozny2002transforming} learns a
non-parametric, piecewise-constant monotone mapping
 
\begin{equation}
    f : [0,1] \to [0,1], \quad f \text{ non-decreasing},
\end{equation}
 
from raw confidence scores to calibrated probabilities.  Given calibration
pairs $\{(p_i, y_i)\}_{i=1}^{N}$ with $p_i \in [0,1]$ and $y_i \in \{0,1\}$,
isotonic regression solves
 
\begin{equation}
    \hat{f} = \arg\min_{f\,\text{non-decreasing}}
              \sum_{i=1}^{N} \bigl(f(p_i) - y_i\bigr)^2,
    \label{eq:isotonic}
\end{equation}
 
via the pool adjacent violators (PAV) algorithm. 
Unlike Temperature Scaling and Beta Calibration, Isotonic Regression
places no parametric assumptions on the shape of the miscalibration, making it
more flexible but also more prone to overfitting on small calibration sets.

In the experiments, we use \textsc{scikit-learn}'s \texttt{IsotonicRegression} with
\texttt{out\_of\_bounds="clip"}, which fits a piecewise-constant,
non-decreasing mapping $f: p \mapsto \hat{p}$ on the flattened validation
scores via the pool adjacent violators algorithm. At inference, the fitted
mapping is applied element-wise, with input values
outside the training range clipped to the nearest boundary.

\subsection{Meta-cal}
Meta-cal~\citep{ma2021meta} is a post-hoc calibration method that combines
temperature scaling with a confidence-aware sample selection strategy.  The key
idea is to avoid calibrating high-uncertainty predictions---which are
unreliable by nature---and instead assign them the uninformative probability
$0.5$.

\paragraph{Step 1 -- Entropy-based partitioning}
For each pixel with pre-sigmoid logit~$z$, the binary entropy of the
uncalibrated prediction $p = \sigma(z)$ is computed as

\begin{equation}
    H(z) = -p\log p - (1-p)\log(1-p).
\end{equation}

A threshold $\tau$ on this entropy separates the predictions into a
\emph{low-entropy} set $\mathcal{L} = \{i : H(z_i) < \tau\}$ and a
\emph{high-entropy} set $\mathcal{H} = \{i : H(z_i) \geq \tau\}$.

\paragraph{Step 2 -- Threshold selection}
The threshold is chosen to match a target coverage accuracy~$a^*$.  A small held-out ($10\%$) of the
calibration data is binned by entropy quantiles, and for each bin the
cumulative accuracy is computed.  Isotonic regression (decreasing) is then
fitted on the (cumulative accuracy, mean entropy) pairs; the threshold is
obtained by evaluating the isotonic fit at the target accuracy:

\begin{equation}
    \tau = \mathrm{IsoReg}^{-1}(a^*).
\end{equation}

\paragraph{Step 3 -- Temperature scaling on low-entropy samples.}
On the remaining calibration data, only samples in~$\mathcal{L}$ are used to
fit a single temperature parameter~$T > 0$ by minimising binary cross-entropy:

\begin{equation}
    T^* = \arg\min_{T > 0}\;
    \sum_{i \in \mathcal{L}}
        \ell_{\mathrm{CE}}\!\bigl(\sigma(z_i / T),\, y_i\bigr).
\end{equation}

\paragraph{Step 4 -- Inference}
At test time, the calibrated probability is

\begin{equation}
    \hat{p} =
    \begin{cases}
        \sigma(z / T^*) & \text{if } H(z) < \tau, \\[4pt]
        0.5             & \text{if } H(z) \geq \tau.
    \end{cases}
\end{equation}

By abstaining from calibrating high-entropy pixels, MetaCal avoids distorting
predictions where the model is genuinely uncertain, which can improve calibration compared to global temperature scaling.
In the experiments $a^*=0.9$.

\subsection{Selective Scaling}
\label{app:selective_scaling}
Selective Scaling~\citep{wang2023calibrating} extends Meta-cal by
learning two temperatures and a binary selector that routes each
prediction to the appropriate one.  The key observation is that correctly
classified and misclassified pixels benefit from different degrees of
rescaling.
\paragraph{Step 1 -- Selector.}
A small MLP $g_\phi$ is trained on $10\%$ of the calibration split to predict
whether the network's hard prediction $\hat{y}_i$ matches the ground-truth
$y_i$.  Its input is the uncalibrated probability
$p_i = \sigma(z_i)$ and its binary target is
$c_i = \mathbf{1}[\hat{y}_i = y_i]$.  After training, the selector partitions
any set of pixels into two groups:

\begin{equation}
    \mathcal{C} = \{i : g_\phi(p_i) = 1\}, \qquad
    \mathcal{M} = \{i : g_\phi(p_i) = 0\}.
\end{equation}
\paragraph{Step 2 -- Two temperatures.}
On the rest of the calibration split, two scalar temperatures $T_1 > 0$
(for predicted-incorrect pixels $\mathcal{M}$) and $T_2 > 0$
(for predicted-correct pixels $\mathcal{C}$) are jointly optimised by
minimising the total cross-entropy:

\begin{equation}
    T_1^*,\, T_2^* = \arg\min_{T_1, T_2 > 0}\;
    \frac{1}{|\mathcal{M}|}\!\sum_{i \in \mathcal{M}}
        \ell_{\mathrm{CE}}\!\bigl(\sigma(z_i / T_1),\, y_i\bigr)
    \;+\;
    \frac{1}{|\mathcal{C}|}\!\sum_{i \in \mathcal{C}}
        \ell_{\mathrm{CE}}\!\bigl(\sigma(z_i / T_2),\, y_i\bigr).
    \label{eq:selective_scaling}
\end{equation}

As in Temperature Scaling, both temperatures are log-parametrised
($T_k = e^{\tau_k}$, $\tau_k \in \mathbb{R}$) to ensure strict positivity,
and the two-parameter problem is solved with L-BFGS.

\paragraph{Step 3 -- Inference.}
At test time, the calibrated probability of a new pixel with logit~$z$ is

\begin{equation}
    \hat{p} =
    \begin{cases}
        \sigma(z / T_1^*) & \text{if } g_\phi\bigl(\sigma(z)\bigr) = 0
            \quad (\text{predicted incorrect}), \\[4pt]
        \sigma(z / T_2^*) & \text{if } g_\phi\bigl(\sigma(z)\bigr) = 1
            \quad (\text{predicted correct}).
    \end{cases}
\end{equation}

In practice, over-confident networks typically yield $T_1^* > 1$ (softening
wrong predictions) and $T_2^* \leq 1$ (sharpening correct ones), producing a
more reliable confidence signal than a single global temperature.

\subsection{Effect of Calibration on Accuracy}
While the main objective of the paper is to compare the calibration performances of different methods, it is natural to ask what is the effect of these calibration methods on the accuracy of the underlying segmentation model. 
To do that we are going to evaluate the DICE score difference, before and after calibration, on the datasets used in the experiments. The DICE score is defined as:
\[DICE=\frac{2|Y\bigcap \hat{Y}|}{|Y|+|\hat{Y}|},\]
where $Y$ and $\hat{Y}$ are the ground-truth and predicted segmentation respectively. The mean is evaluated across random calibration/test splits.
\begin{table}
	\caption{Relative DICE score variation after calibration.}  

	\label{tab:pre_post_cal} 
	\begin{center} 
	\begin{tabular}{lrrrr}  
		\toprule
		\textbf{Methods} & \textbf{Cityscapes} & \textbf{COCO-an.} & \textbf{COCO-veh.} & \textbf{TN3K}\\  
		\midrule  
		Beta-cal & $2.30\%$ & $0.22\%$& $0.33\%$&$0.05\%$   \\  
		Iso.Regr.  & $1.55\%$ & $0.36\%$& $0.32\%$&$-0.55\%$  \\ 
    Meta-cal  & $-33.59\%$ & $-14.04\%$& $-8.28\%$&$-53.04\%$  \\ 
    Logistic  & $2.6\%$ & $0.36\%$& $0.33\%$&$-0.02\%$  \\  
		Binning (B=10) & $-1.51\%$ & $0.05\%$ &$-0.07\%$& $-3.43\%$  \\
		Binning (B=30) &  $1.49\%$  & $0.36\%$ & $0.22\%$& $-14.2\%$ \\  
		\bottomrule  
	\end{tabular}   
    \end{center}
 
\end{table} 
\begin{table}
	\caption{Relative DICE score variation after calibration.}  
	\label{tab:pre_post_cal2}   
	\begin{center}  
	\begin{tabular}{lrrrrr}  
		\toprule
		\textbf{Methods} & \textbf{ISIC} & \textbf{Kvasir} & \textbf{BUSI} & \textbf{FLAIR1} & \textbf{Pet}\\  
		\midrule  
		Beta-cal  & $-0.86\%$ & $-0.33\%$& $7.11\%$&$-0.91\%$& $-0.01\%$   \\  
		Iso.Regr.  & $-0.19\%$ & $-0.75\%$& $8.31\%$&$-0.56\%$& $-0.01\%$ \\  
    Meta-cal  & $-26.30\%$ & $-43.60\%$& $-0.37\%$&$-2.89\%$ & $-30.63\%$\\ 
    Logistic  & $-0.41\%$ & $-0.07\%$& $4.21\%$&$-3.34\%$ & $-0.04\%$\\  
		Binning (B=10)& $-0.33\%$ & $-9.11\%$& $2.56\%$&$-0.20\%$& $-0.53\%$  \\
		Binning (B=30)& $-0.49\%$ & $-0.87\%$& $3.84\%$&$-0.67\%$& $-0.38\%$ \\  
		\bottomrule  
	\end{tabular} 
  \end{center} 
\end{table} 
Tables \ref{tab:pre_post_cal}--\ref{tab:pre_post_cal2} show that \textit{post-hoc} calibration methods yield, on average, a marginal decrease in segmentation accuracy on the order of a tenth of a percentage point. 
There are some exceptions, such as Cityscapes or BUSI, where the calibration brings also an improvement in the accuracy. 
Meta-cal is the worst method concerning the effect on DICE, often substantially decreasing the accuracy.
Moreover, we observe that a larger number of bins is generally preferable: too few bins produce overly coarse intervals, increasing the risk of incorrectly reassigning the confidence values of individual pixels. 
Temperature scaling and Selective scaling do not change the accuracy of the model, hence are not discussed.
 
 
 
\section{Datasets and Pre-trained Models}
\label{app:datasets_and_models}

\subsection{Datasets}

\paragraph{Cityscapes}~\citep{Cordts2016Cityscapes,Cordts2015Cvprw} is a large-scale benchmark for semantic understanding of urban street scenes, collected from a moving vehicle across 50 cities under varying seasonal and lighting conditions. It provides 5{,}000 images with high-quality pixel-level annotations and an additional 20{,}000 coarsely annotated frames. The fine-annotation split is pre-divided into 2{,}975 training, 500 validation, and 1{,}525 test images at a resolution of $1024 \times 2048$ pixels. Annotations span 30 semantic classes organized into 8 higher-level categories, covering both static elements (road, building, vegetation) and dynamic actors (person, car, cyclist). Its combination of high resolution, class diversity, and instance-level masks has made it the de-facto benchmark for segmentation in autonomous driving. 
 
\paragraph{MS COCO}~\citep{lin2014microsoft} (Common Objects in Context) is a large-scale, general-purpose dataset for object detection, instance segmentation, and image captioning. It contains over 330{,}000 images with more than 2.5 million labeled instances drawn from 80 foreground object categories and 91 background 'stuff' categories. Its scale and annotation richness have established COCO as the primary standard for evaluating general-purpose segmentation models. We have created two subsets from the original dataset: COCO- animals and COCO-vehicles. Specifically, we selected images that feature either animals or humans to form the COCO-animals dataset, and images that contain vehicles or humans to create the COCO-vehicles dataset. 
The COCO-animals dataset includes 314 images with the following labels: background, cat, dog, sheep, cow, horse, bird, and human. 
In contrast, the COCO-vehicles dataset contains 321 images with these labels: background, train, bus, bicycle, airplane, car, boat, and human. We used those images for the calibration experiments. 
 
\paragraph{TN3K}~\citep{GONG2023106389} The TN3K (Thyroid Nodule Region Segmentation Dataset) comprises 3493 ultrasound images from 2421 patients, captured between January 2016 and August 2020. These images were selected from over 30,000 images provided by different hospitals.

\paragraph{ISIC 2016}~\citep{gutman2016skin} is the first public benchmark released by the International Skin Imaging Collaboration (ISIC) for automated skin lesion analysis, hosted as part of the ISBI 2016 challenge. It contains 900 dermoscopic training images and 379 test images, each annotated with a binary pixel-level mask delineating the lesion boundary against the surrounding skin. The classification task is binary (melanoma vs.\ benign), and the segmentation ground-truth was produced by expert clinicians.

\paragraph{Kvasir-SEG}~\citep{Pogorelov:2017:KMI:3083187.3083212} is an open-access dataset of 1{,}000 gastrointestinal polyp images captured during colonoscopy, each paired with a manually delineated binary segmentation mask and a corresponding bounding box annotation. Masks were initially annotated by a medical doctor and then verified by an experienced gastroenterologist, ensuring high ground-truth quality. Accurate polyp segmentation is clinically important because polyps are precursors to colorectal cancer. 

\paragraph{BUSI}~\citep{tasnim2024cam, al2020dataset} is a publicly available collection of breast ultrasound images. It contains 780 images from 600 female patients of age ranging between 25 and 75 years, featuring malignant and benign breast cancers but also normal cases without cancer. The dataset had few anomalies that were later corrected.
 
\paragraph{FLAIR1}~\citep{ign2022flair1} The FLAIR \#1 (French Land cover from Aerospace ImageRy) dataset is a large-scale, high-resolution, aerial imagery dataset designed for semantic segmentation of land cover across France. Developed by the French National Institute of Geographical and Forest Information (IGN), it aims to address challenges in domain adaptation, such as environmental heterogeneity and temporal changes. The FLAIR datasets are released under the Open Licence 2.0 from Etalab.
 
\paragraph{Oxford-IIIT Pet}~\citep{parkhi12a} is a fine-grained recognition and segmentation dataset comprising approximately 7390 images of cats and dogs spanning 37 breeds, with roughly 200 images per class. 

\subsection{Pre-trained Models and Data Splits}

\paragraph{U-Net}
We train a lightweight U-Net~\cite{ronneberger2015u} from scratch for
binary foreground segmentation. The encoder follows a standard contracting
path with four downsampling stages: starting from a base width $b = 32$,
each stage doubles the number of channels ($b \to 2b \to 4b \to 8b \to 16b$,
i.e.\ up to 512 feature maps at the bottleneck) and halves the spatial
resolution via $2{\times}2$ max-pooling. Every resolution level applies two
successive $3{\times}3$ convolutions, each followed by batch normalisation
and ReLU activation; a spatial dropout layer ($p = 0.2$) is inserted between
the two convolutions to regularise training. The decoder mirrors the encoder
with four transposed-convolution upsampling steps ($2{\times}2$, stride 2),
each concatenating the corresponding encoder skip connection before applying
the same double-convolution block. A final $1{\times}1$ convolution projects
the $b$-channel feature map to a single output logit per pixel; the sigmoid
activation is applied externally. The model takes $3$-channel RGB images of
arbitrary spatial resolution as input and produces raw logits of the same
spatial size.

This architecture has been used for the following datasets: ISIC 2016, BUSI, Kvasir-SEG and TN3K. Each training has been performed with $200$ epochs, batch size of $8$ and learning rate of $10^{-3}$. Images and
masks are resized to $256{\times}256$ pixels; images are normalized with
ImageNet mean and standard deviation.
For ISIC, the dataset is already divided in train/val/test. We use the train and val splits to train the U-Net and  the test split, featuring 379 images, for the calibration experiments. 
For BUSI, we use only the images with foreground (cancer). We train the U-Net with 439 images and use only 150 for the calibration experiments. 
This dataset is the smallest one in our evaluation and shows that meaningful calibration can be achieved even with few images. 
For Kvasir, we use 700 images to train the U-Net and the remaining 300 for the calibration experiments. 
Finally for TN3K, we
train the U-Net with 2879 images and used the remaining 614 for the calibration experiments. 

\paragraph{DeepLabV3 Fine-tuning}
For the Oxford-IIIT Pet dataset we fine-tune a
deeplabv3\_resnet50 model from \texttt{torchvision.models}. The
original classifier head (21-class) is replaced by a $1{\times}1$
convolutional layer mapping the 256-channel output to 2 classes
(foreground / background). 

The dataset consists of JPEG images paired
with annotations; the annotations are binarized so that label $1$ (pet) maps
to the foreground class and all other values map to background. Images and
masks are resized to $224{\times}224$ pixels; images are normalized with
ImageNet mean and standard deviation. A 90/10 random split
yields training and test subsets. The model is optimized with
Adam ($\text{lr} = 10^{-4}$) using a pixel-wise cross-entropy loss over
mini-batches of 32 images for 5 epochs. 
We use 6651 images for training and obtain the prediction of 739 images for the calibration experiments.

\paragraph{ResNet34-UNet}
For the FLAIR1 challenge we use a pre-trained model (\textit{FLAIR-INC\_rgbie\_15cl\_resnet34-unet}) provided by the organizer\footnote{\url{https://ignf.github.io/FLAIR/}}.
We use the pre-trained model on the smallest test subsets (\textit{D022\_2021}) containing only 500 images, which are later used in our calibration experiments.
The predictions contain 19 different classes which are binarized with labels=$0, 8, 10, 11, 12, 17$ as foreground and the rest as background.

\paragraph{DeepLabV3+ Models} 
For the Cityscapes dataset we use a DeepLabV3Plus-ResNet101\footnote{\url{https://github.com/VainF/DeepLabV3Plus-Pytorch}} backbone \cite{chen2018encoder,chen2019rethinking} on the validation set featuring 500 images. 
All of the predictions were randomly cropped to $256\times256$ and binarized with labels=$11,\dots,18$ as foreground and the rest as background. 
For the COCO-animals experiment we use the DeepLabv3plus\_ResNet50 backbone,  while for the COCO-vehicles the DeepLabv3plus\_MobileNet backbone. 
All of the predictions were binarized with labels=$1,\dots,7$ as foreground and the rest as background. The code is published under the MIT License.

\paragraph{Data Splits}
During the experiments we perform $K$ random splits of the data. 
For each random split we divide our dataset $\mathcal{D}$, in three subsets $\mathcal{D}_{bin},\ \mathcal{D}_{cal}$ and $\mathcal{D}_{test}$.
For the \textit{post-hoc} baselines, $\mathcal{D}_{bin}$ and $\mathcal{D}_{cal}$ are used to calibrated the models.
For the binning approach, $\mathcal{D}_{bin}$ is used to construct the bins, and $\mathcal{D}_{cal}$ is used to construct the calibrator $H$.

\begin{table}
\centering
\caption{Comparison of post-hoc calibration baselines.}
\label{tab:dataset_comparison}
\begin{tabular}{lcc}
\hline
\textbf{Dataset} & \textbf{Pre-trained model}  &
\textbf{\# samples for exp. (cal.+test)}  \\
\hline
Cityscapes & DeepLabv3plus\_ResNet101  & 500 \\
COCO-animals & DeepLabv3plus\_ResNet50  & 314 \\
COCO-vehicles & DeepLabv3plus\_MobileNet  & 321 \\
TN3K & U-Net  & 614 \\
ISIC & U-Net  & 379 \\
Kvasir & U-Net  & 300 \\
BUSI & U-Net  & 150 \\
FLAIR1 & ResNet34-UNet  & 500 \\
Oxford-Pet & DeepLabv3\_resnet50 + fine tuning  & 739 \\
\hline
\end{tabular}
\end{table}

\section{Theoretical Bound Estimation}

\label{app:appendix_bound_estimation}
\subsection{$\bar D$ estimation with autoregressive model} 
\label{app:Dbarestimation}
We now go through the implementation of the autoregressive model used to estimate the matrix $\bar D.$
The coupling matrix $D^\sigma_{x,y}$ quantifies how much the distribution
of $\sigma_y$ changes when $\sigma_x$ is intervened upon, holding the past
$\sigma_{<x}$ fixed.  Evaluating this quantity requires access to the
full conditional distribution $\mathbb{P}(\sigma_y \mid \sigma_{\le x})$, which in
turn depends on the joint distribution over the mask.
The autoregressive factorisation,
\[
  \mathbb{P}_\theta(\sigma) = \prod_{x\in\Omega}
    \mathbb{P}_\theta(\sigma_x \mid \sigma_{<x}),
\]
expresses the conditional $\mathbb{P}(\sigma_y \mid \sigma_{\le x})$ as a marginal
over the intermediate sites $\sigma_{x+1},\dots,\sigma_{y-1}$. 
This marginal is approximated by Monte Carlo.
Moreover, since in our application the variable $\sigma_x$ are conditioned on a realization of the binning scheme and since we need to consider only the labels in a specific bin $b$, 
we construct the autoregressive model such that it takes as input a partition $\Omega_b$ and calibrated predictions $h(X)$, resulting in the factorisation:
\begin{equation}
  \label{eq:cond_bin}
  \mathbb{P}_\theta(\sigma\mid h(X)) = \prod_{x_b\in\Omega_b}
    \mathbb{P}_\theta(\sigma_{x_b} \mid \sigma_{<x_b},h(X)_{x_b}),
\end{equation}
where $\sigma \in \{0,1\}^{\Omega_b}$ is a binary mask and $\Omega_b$ is a subset of a 2-D grid $\Omega$.
\paragraph{Model architecture}
To ease the notation we are going to drop the $b$-dependent notation.
We model the per-site conditional as a PixelCNN \cite{van2016pixel,van2016conditional}
\begin{equation}\label{eq:cond}
  \mathbb{P}_\theta\!\left(\sigma_x = 1 \middle| \sigma_{<x}, h(X)_x\right)
\end{equation}
with sites in raster order, and ${\sigma_{<x} = (\sigma_1, \dots, \sigma_{x-1})}$.
The network has two parallel streams: a
$\sigma$-stream of causally masked $k \times k$ convolutions carrying
$\sigma_{<x}$, and an $h$-stream of $1 \times 1$ convolutions only,
carrying $h(X)_x$. The streams are summed at the readout,
so~\eqref{eq:cond} depends on $\sigma_{<x}$ and on $h(X)_x$ alone and never
on $h(X)_y$ for $y \neq x$.

For a kernel of odd size $k$ with center at $c = (k{-}1)/2$, the causal
masks are
\begin{equation}
  M^{(\mathrm{A})}_{x_1,x_2}
  = \mathbf{1}\!\big[x_1 < c \;\text{or}\; (x_1 = c,\, x_2 < c)\big],
  \qquad
  M^{(\mathrm{B})}_{x_1,x_2}
  = M^{(\mathrm{A})}_{x_1,x_2} + \mathbf{1}\!\big[(x_1,x_2) = (c,c)\big].
\end{equation}
Mask~A excludes the center, mask~B includes it. A masked convolution
multiplies the weight tensor by $M^{(\tau)}$ before convolving:
$\mathrm{MaskedConv}^{(\tau)}_{k}(w) = (W \odot M^{(\tau)}) * w + b$.

The $\sigma$-stream uses mask~A in its first layer (so its output at site
$x$ does not depend on $\sigma_x$) and mask~B thereafter (centre allowed
because upstream activations are already strictly causal in $\sigma$):
\begin{align}
  s^{(0)} &= \mathrm{MaskedConv}^{(\mathrm{A})}_{k}(\sigma),\\[2pt]
  s^{(\ell+1)} &= s^{(\ell)}
  + \mathrm{MaskedConv}^{(\mathrm{B})}_{k}\!\big(
      ReLU\!\big(
        \mathrm{MaskedConv}^{(\mathrm{B})}_{k}\!\big(
          ReLU(s^{(\ell)})
        \big)
      \big)
    \big),
  \qquad \ell = 0,\dots,L{-}2.
\end{align}
The $h$-stream has the same skeleton with $k=1$ and no masking:
$c^{(0)} = Conv_1(h(X)_x)$ and residual blocks formed analogously from
two $1 \times 1$ convolutions and a $ReLU$.

The readout sums the streams and applies a final pointwise layer:
\begin{equation}
  \mathbb{P}_\theta\!\left(\sigma_x = 1 \,\middle|\, \sigma_{<x},\, h(X)_x\right)
  = \operatorname{sigmoid}\!\big(Conv_1\!\big(
      ReLU\!\big(s^{(L-1)} + c^{(L-1)}\big)
    \big)_x\big).
\end{equation}


\paragraph{Coupled sampling}

Concretely, we estimate $D^\sigma_{x,y}$ by sampling with a
common-random-numbers coupling \citep{mohamed2020monte,glasserman2004monte}. Fix a context $\sigma_{<x}$, then for each Monte Carlo trial we run two chains
$A$ and $B$:
\[
  \sigma^A_x = 1, \qquad \sigma^B_x = 0,
  \qquad \sigma^A_{<x} = \sigma^B_{<x} = \sigma_{<x}.
\]
For each subsequent site $t = x{+}1, \dots, HW$ (where $H\times W$ are the image dimensions) we perform one batched
forward pass to obtain the two conditional probabilities
\[
  p^A_t = P_\theta(\sigma_t = 1 \mid \sigma^A_{<t}),
  \qquad
  p^B_t = P_\theta(\sigma_t = 1 \mid \sigma^B_{<t}),
\]
draw a {single} threshold $u_t \sim \mathcal{U}[0,1]$ shared between
the two chains, and set
\[
  \sigma^A_t = \mathbf{1}_{[u_t < p^A_t]},
  \qquad
  \sigma^B_t = \mathbf{1}_{[u_t < p^B_t]}.
\]
Each chain therefore advances under its own conditional --- so neither
estimator is biased --- while the shared threshold ensures that sites
where the two conditionals agree contribute exactly zero to the
difference $\sigma^A_t - \sigma^B_t$.  Averaged over $M$ trials,
\[
  \hat{D}^\sigma_{x,t}(\sigma_{<x})
    \;=\; \Bigl|\, \frac{1}{M}\sum_{m=1}^{M}
      \bigl(\sigma^{A,(m)}_t - \sigma^{B,(m)}_t\bigr) \Bigr|
\]
is an unbiased Monte Carlo estimator of $|D^\sigma_{x,t}|$ conditional
on $\sigma_{<x}$, with variance reduced compared to two
independent runs.

\paragraph{Full decay curve}

A key observation is that the autoregressive structure allows us to
estimate $D^\sigma_{x,y}$ for all $y > x$ simultaneously from a
single pair of coupled trajectories.  The recorded differences
$\sigma^A_t - \sigma^B_t$ at every $t > x$ already provide one Monte
Carlo realisation of $D^\sigma_{x,t}$ at every target site.  Repeating
the coupled run $M$ times under the same $\sigma_{<x}$ gives the full
empirical decay $\hat{D}^\sigma_{x,t}(\sigma_{<x})$ for $t = x{+}1,
\dots, HW$. 




\paragraph{Aggregation and the decay fit}

To reduce dependence on any particular grid location, we repeat the
procedure for $n_s$ source sites $x$ drawn from the interior of the
grid,
and for $n_c$ test contexts providing different prefixes $\sigma_{<x}$.
For each source--context pair we compute the per-target estimates
$\hat{D}^\sigma_{x,y}(\sigma_{<x})$ as described above and bin them
by Euclidean distance $d = \|y - x\|_2$.
The aggregation across the three nested sources of randomness ---
trials, contexts, sources --- is performed as follows:
\begin{enumerate}
  \item Within each $(x, \sigma_{<x})$, the Monte Carlo trials are
        averaged ($\hat{D}^\sigma_{x,y}(\sigma_{<x})$).
  \item For each context, we average the per-target values over all
        $y$ falling in the same distance bin $d$, yielding one number
        per $(x, \sigma_{<x}, d)$.
  \item Across contexts, we take the {maximum} over $\sigma_{<x}$.
        This is a worst-case aggregation: the calibration bound of
        Theorem~\ref{thm:improved} depends on $\sup_{\sigma_{<x}}
        |D^\sigma_{x,y}|$ rather than its average, since the bound
        must hold for every realisable prefix.
  \item Across sources, we average the resulting per-source worst-case
        values to obtain the empirical decay curve $\hat{D}(d)$.
\end{enumerate}
Fitting the parametric form~\eqref{eq:fit} to this curve serves two
purposes.  First, it provides a closed-form approximation of
the coupling matrix for any distance, including distances not
directly observed in the finite grid. 
Second, it allow us to compute the coefficients $S$ and $c$ of the bounds.

Moreover, since we need to estimate the coupling matrix for variables $Y_{x_b}$ in bin $b$, we have to use the factorisation in \eqref{eq:cond_bin}.
In practice, we estimate the matrix $\hat D(d)$ for different bins $b$ and average the results.

\paragraph{Experiment settings}
During the experiments we use the calibration dataset ($\mathcal{D}_{cal}$) to estimate the matrix $\hat{D}$.
We sample random contexts, of shape $32\times 32$ ($64\times 64$ for the FLAIR dataset, due to the larger image dimensions) from the ground-truth segmentation.
We use $5000$ contexts to train the autoregressive model. We use the Adam optimizer and train the model for $100$ epochs with learning rate $10^{-3}$ and batch size equal to $32$.

For the evaluation the hyperparameters are: the number of random source points $n_s = {20}$, the number of random contexts $n_c = {20000}$ and the number of random Monte Carlo trials $M = {10}$.
We use masks of size $3\times 3$, with 32 hidden channels and 1 layer.

\subsection{Bound estimation}
Now we can compute the final bound.
The row sums of the global coupling matrix are defined as
\[
  R_{x_b} \;=\; \sum_{\changes{y_b\geq x_b}}
            \hat{D}\!\bigl(\norm{y_b - x_b}_2\bigr),
\]
where $\hat{D}(d) = \hat a\, \exp(-\hat\alpha d^{\hat b})$ is the fitted decay, $\|y - x\|_2$ is the Euclidean distance between pixels
$x$ and $y$, and $y > x$ refers to the raster order of the
autoregressive scan.  Intuitively, $R_{x_b}$ aggregates the total residual
dependence between $x_b$ and all in-bin pixels that follow it in the
scan. From these row sums we can build the coefficients of Theorem~\ref{thm:improved}:
\[
  S_b^{(i)} \;=\; \frac{N_{\mathrm{cal}}}{N_b^{2}}\changes{\hat{q}} \sum_{x_b \in \Omega^{(i)}_b} (R^{(i)}_{x_b})^2,
  \qquad
  c_b^{(i)} \;=\; \frac{1}{3 N_b}\,\max_{x_b \in \Omega^{(i)}_b} R^{(i)}_{x_b},
\]
where \changes{$\hat{q}$ is the maximum variance across contexts estimated
in~\ref{app:Dbarestimation}} and $N_b = \sum_{i=1}^{N_{cal}} |\Omega^{(i)}_b|$ is the
total number of bin-$b$ pixels across $\mathcal{D}_{\mathrm{cal}}$.
In this estimation we assume for simplicity that all the images contributes the same.
We approximate the bound by averaging over a
random subset ($n=20$) of calibration images,
\[
  \bar S_b \;=\; \tfrac{1}{n}\sum^{n}_{i=1} S_b^{(i)},
  \qquad
  \bar c_b \;=\; \tfrac{1}{n}\sum^{n}_{i=1} c_b^{(i)}.
\]
Moreover, for the marginal experiments we use an average of the coefficients:
\[
  \bar S \;=\; \tfrac{1}{B}\sum_{b=1}^{B} \bar S_b,
  \qquad
  \bar c \;=\; \tfrac{1}{B}\sum_{b=1}^{B} \bar c_b.
\]

\paragraph{Final bound}

Assembling all components, the marginal calibration bound (used in the experiments) at tolerance
$\varepsilon$ takes the form of
Theorem~\ref{thm:improved}:
\[
  \mathbb{P}\!\Bigl(\,
    |h(X)_x- \mathbb{E}[Y_x\mid h(X)_x]|\geq \varepsilon\,
  \Bigr)
  \leq 2\exp\bigg(-\frac{\bar S}{\bar c^{2}}\eta(u)\bigg),
  \qquad
  u = \frac{\bar c \varepsilon}{\bar S},
  \qquad
  \eta(u) = 1 + u - \sqrt{1 + 2u}.
\]
The conditional bound is instead approximated as
\[
  \mathbb{P}\!\Bigl(\forall r\in Range(h):
    |r- \mathbb{E}[Y_x\mid h(X)_x=r]| \geq \varepsilon\,
  \Bigr)
  \leq 2\sum^B_{b=1}\exp\bigg(-\frac{\bar S_b}{\bar c_b^{2}}\eta(u_b)\bigg).
\]

\section{Implementation details and Computational Expenses}
\label{app:Implementation_details}
We perform each experiment using a GPU NVIDIA A100-SXM4-40GB. The average computing time of each method, for a calibration split, is given in Table \ref{tab:times}.
\begin{table}
	\caption{Computational time of the calibration methods.}  
	\label{tab:times} 
  \begin{center}
	\begin{tabular}{lccccccc}  
		\toprule
		 & \textbf{Binning} & \textbf{T-scal.} & \textbf{Beta-cal} & \textbf{Log.-cal} & \textbf{Iso-regr} & \textbf{Meta-cal} & \textbf{Sel.Scal}\\  
		\midrule  
		\textbf{Time (s)} & $0.6-5$ & $6-55$& $12-45$ & $4-11$& $3-21$& $6-59$ & $200-1200$   \\   
		\bottomrule  
	\end{tabular} 
  \end{center}   
\end{table} 

\section{Additional Experiments}
\label{app:addexp}
\subsection{Additional Calibratio Experiments}
\begin{figure}
    \centering
    \includegraphics[width=0.9\linewidth]{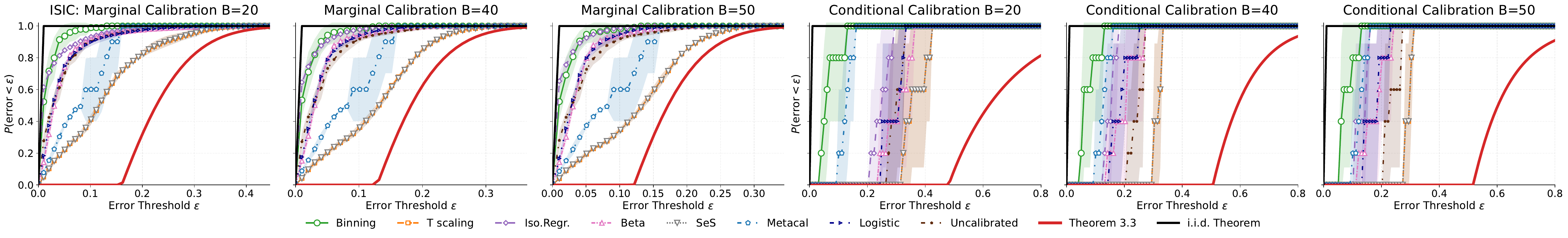}
    \includegraphics[width=0.9\linewidth]{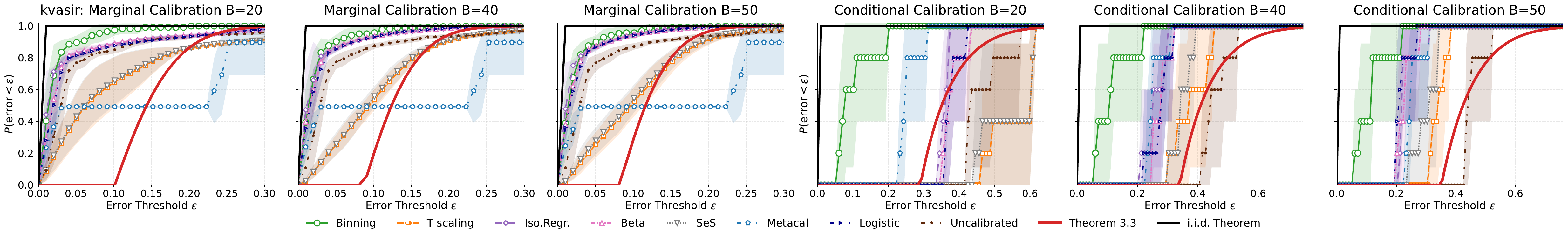}
    \includegraphics[width=0.9\linewidth]{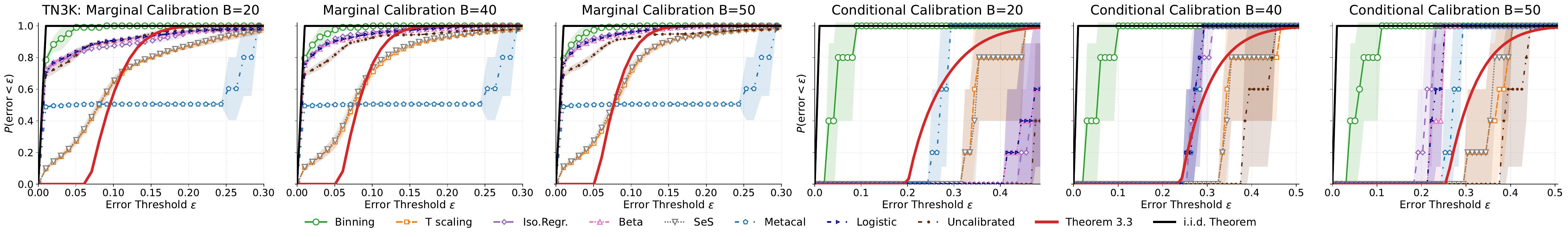}
    \includegraphics[width=0.9\linewidth]{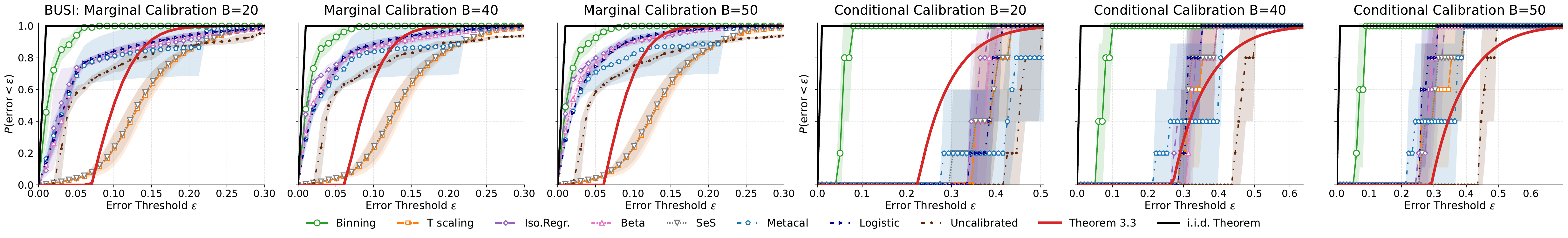}
    \includegraphics[width=0.9\linewidth]{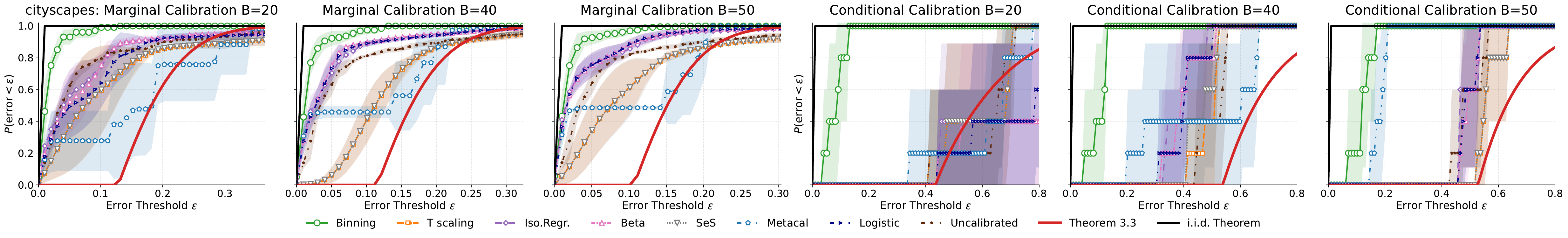}
    \includegraphics[width=0.9\linewidth]{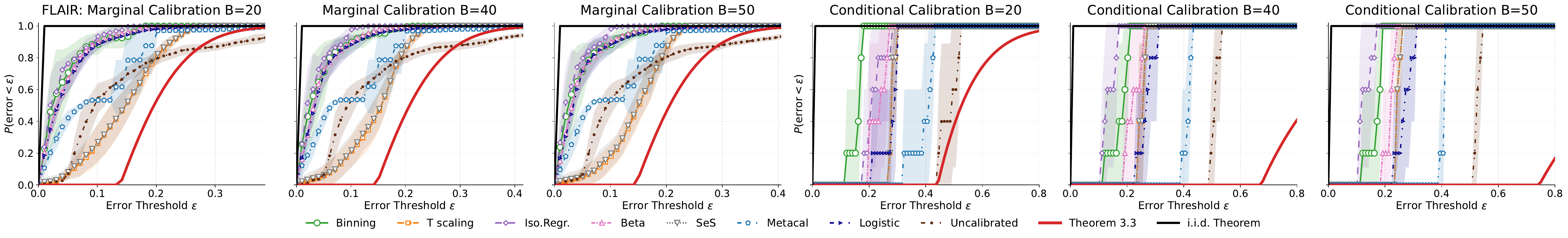}
    \includegraphics[width=0.9\linewidth]{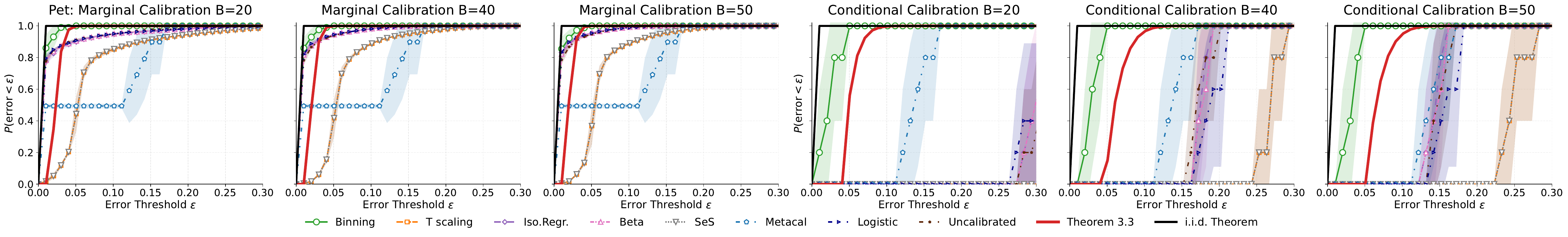}
    \includegraphics[width=0.9\linewidth]{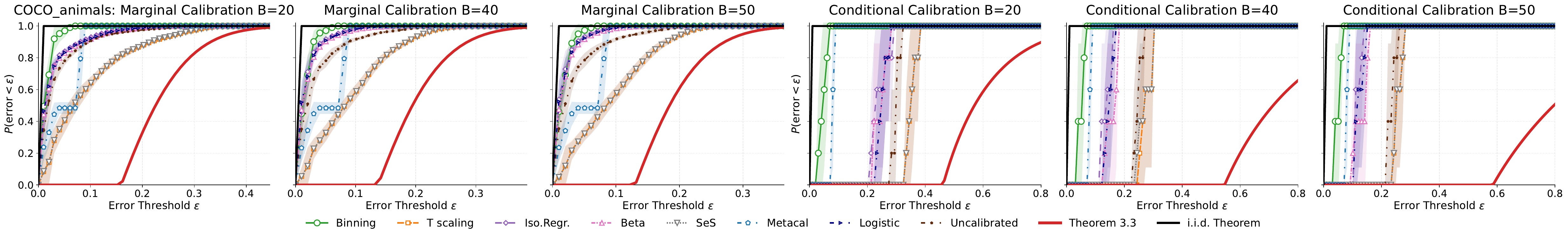}
    \includegraphics[width=0.9\linewidth]{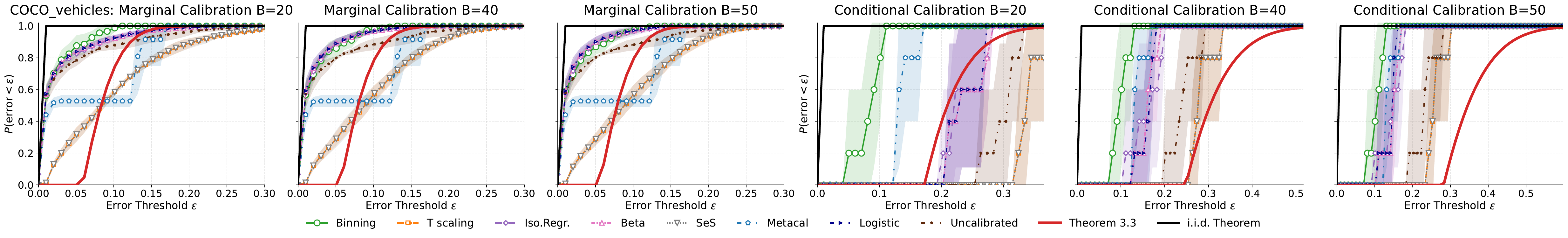}
    \caption{Additional comparisons of validity curves ${V}_m(\varepsilon), {V}_c(\varepsilon)$ (in red) and empirical ones $\hat{V}_m(\varepsilon), \hat{V}_c(\varepsilon)$ across methods and datasets.}
    \label{fig:addresults}
\end{figure}

\begin{figure}
    \centering
    \includegraphics[width=0.9\linewidth]{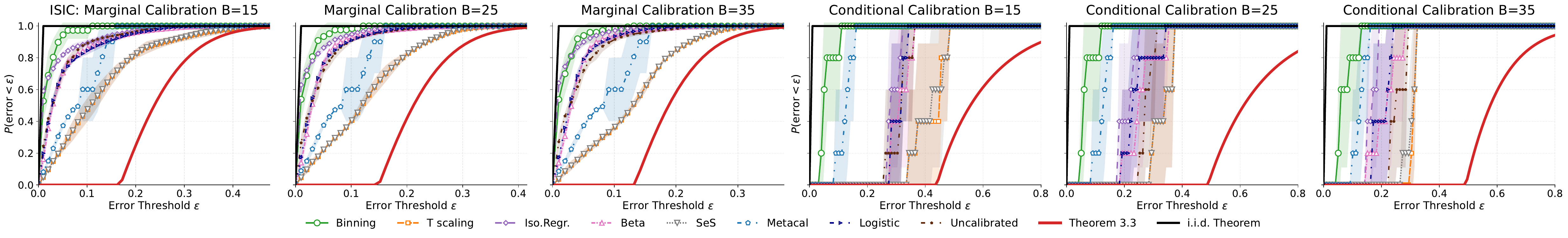}
    \includegraphics[width=0.9\linewidth]{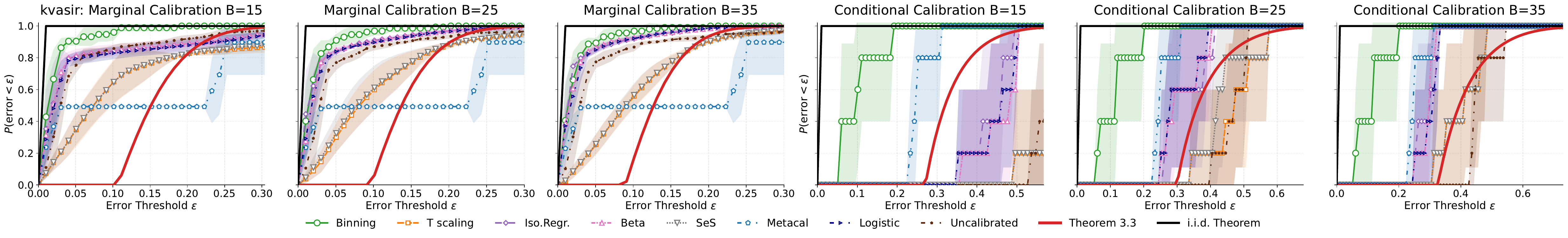}
    \includegraphics[width=0.9\linewidth]{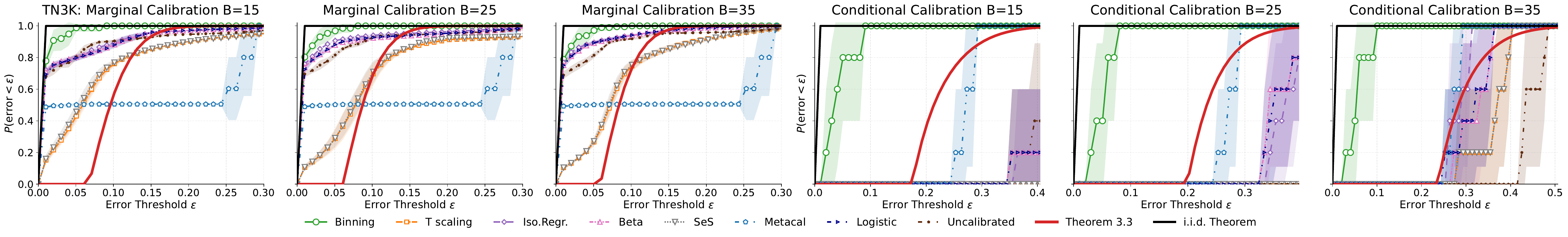}
    \includegraphics[width=0.9\linewidth]{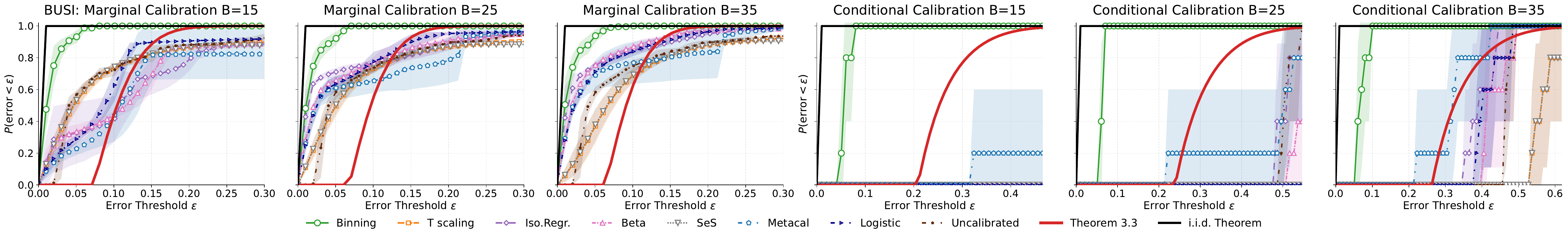}
    \includegraphics[width=0.9\linewidth]{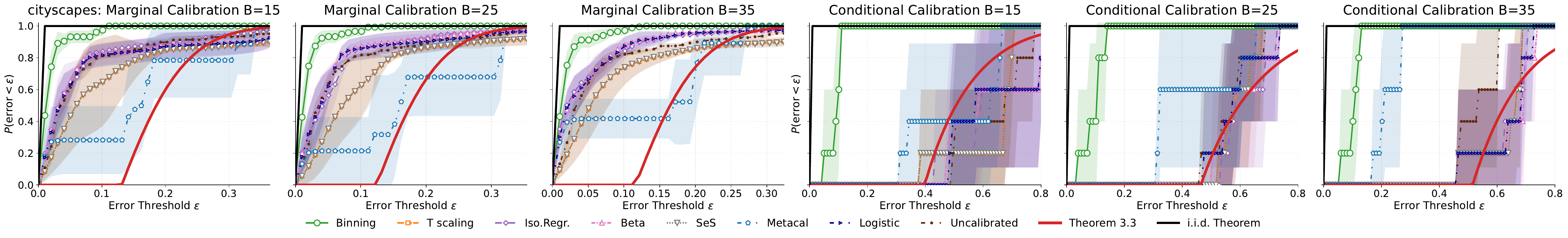}
    \includegraphics[width=0.9\linewidth]{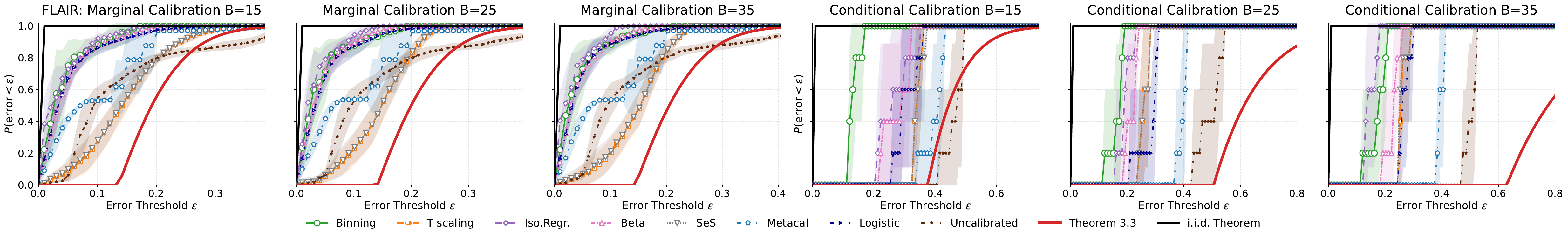}
    \includegraphics[width=0.9\linewidth]{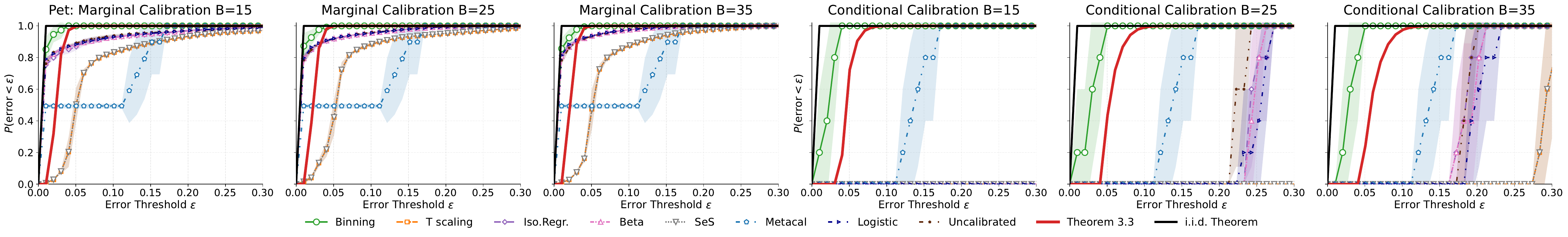}
    \includegraphics[width=0.9\linewidth]{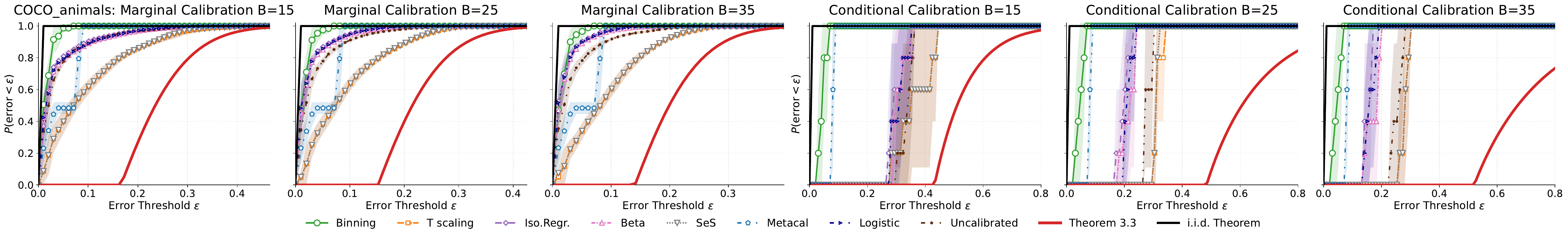}
    \includegraphics[width=0.9\linewidth]{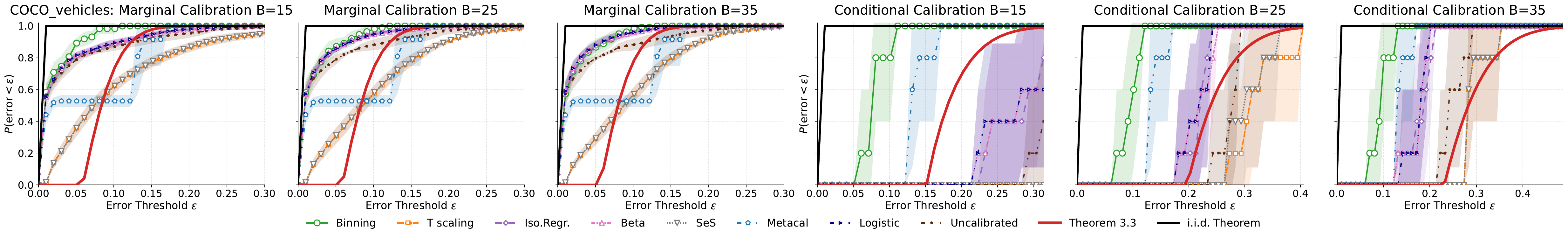}
    \caption{Additional comparisons of validity curves ${V}_m(\varepsilon), {V}_c(\varepsilon)$ (in red) and empirical ones $\hat{V}_m(\varepsilon), \hat{V}_c(\varepsilon)$ across methods and datasets.}
    \label{fig:addresults2}
\end{figure}

\changes{In Figures~\ref{fig:addresults}-\ref{fig:addresults2} we report additional experiments for different choices of $B$.}

\subsection{Binning Conditional Independece Assumption of Theorem \ref{marandcon_cal}}
\changes{In this section we discuss the validity of the binning conditional independence assumption of Theorem~\ref{marandcon_cal}, which states that $Y_x\mid \tilde{h}(X)_x\perp\!\!\!\perp \tilde{h}(X)_y,\ y\neq x$.
In words, the assumption requires that, once the bin assignment of pixel $x$ is known, the bin assignments of the neighboring pixels carry no additional information about the label at $x$.
The intuition behind it is that the segmentation model $f$ is trained to aggregate information from surrounding pixels, so $f(X)_x$---and hence $h(X)_x$---has already absorbed the neighborhood context.
This does not imply conditional independence, but it may make it hold approximately.}

\begin{figure}
    \centering
    \includegraphics[width=0.24\linewidth]{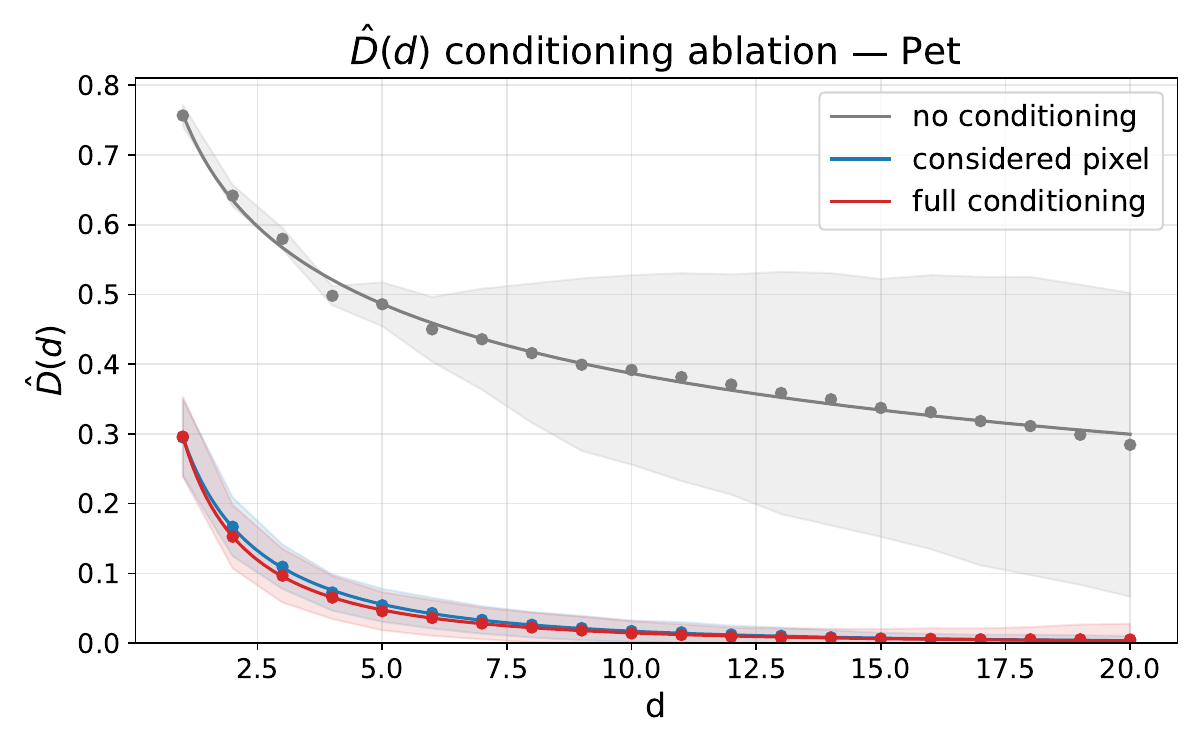}
    \includegraphics[width=0.24\linewidth]{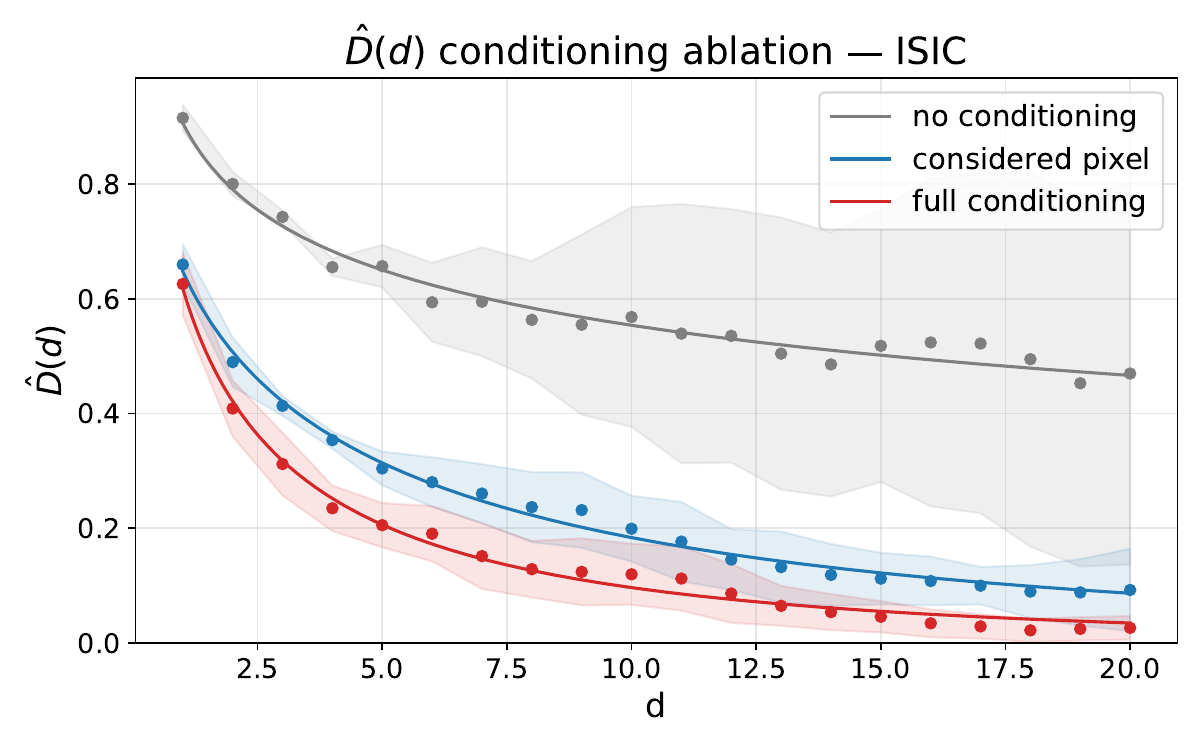}
    \includegraphics[width=0.24\linewidth]{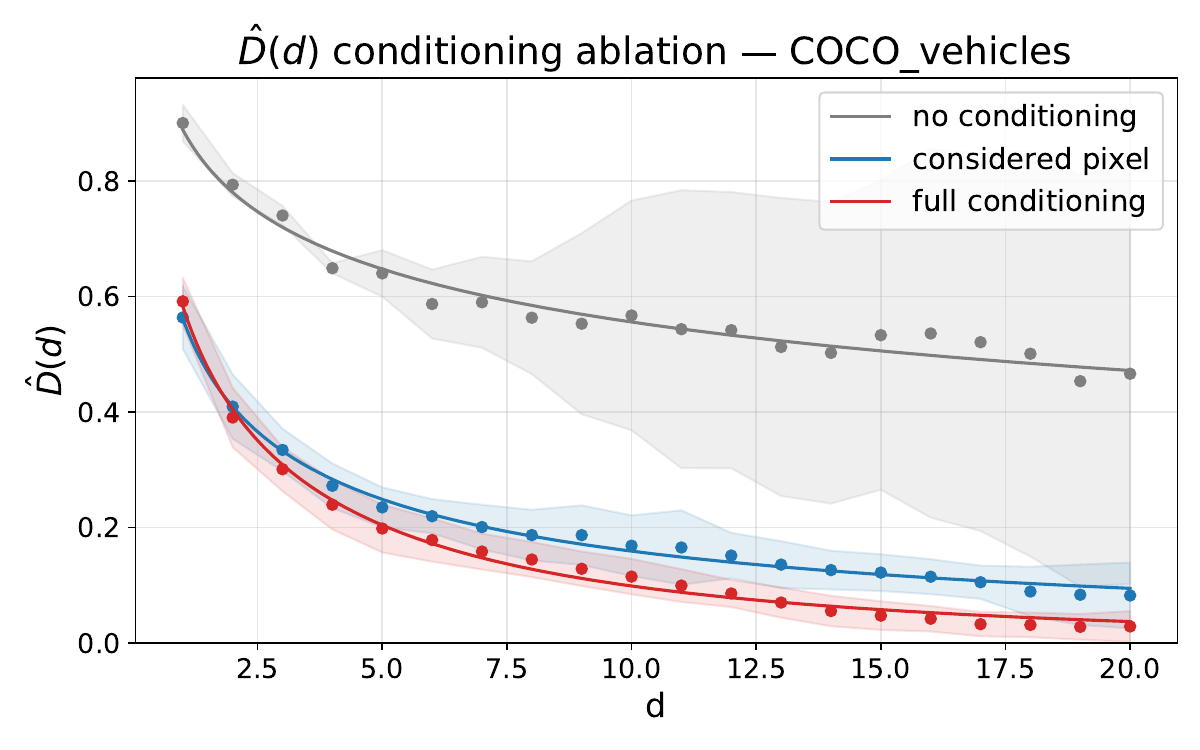}
    \includegraphics[width=0.24\linewidth]{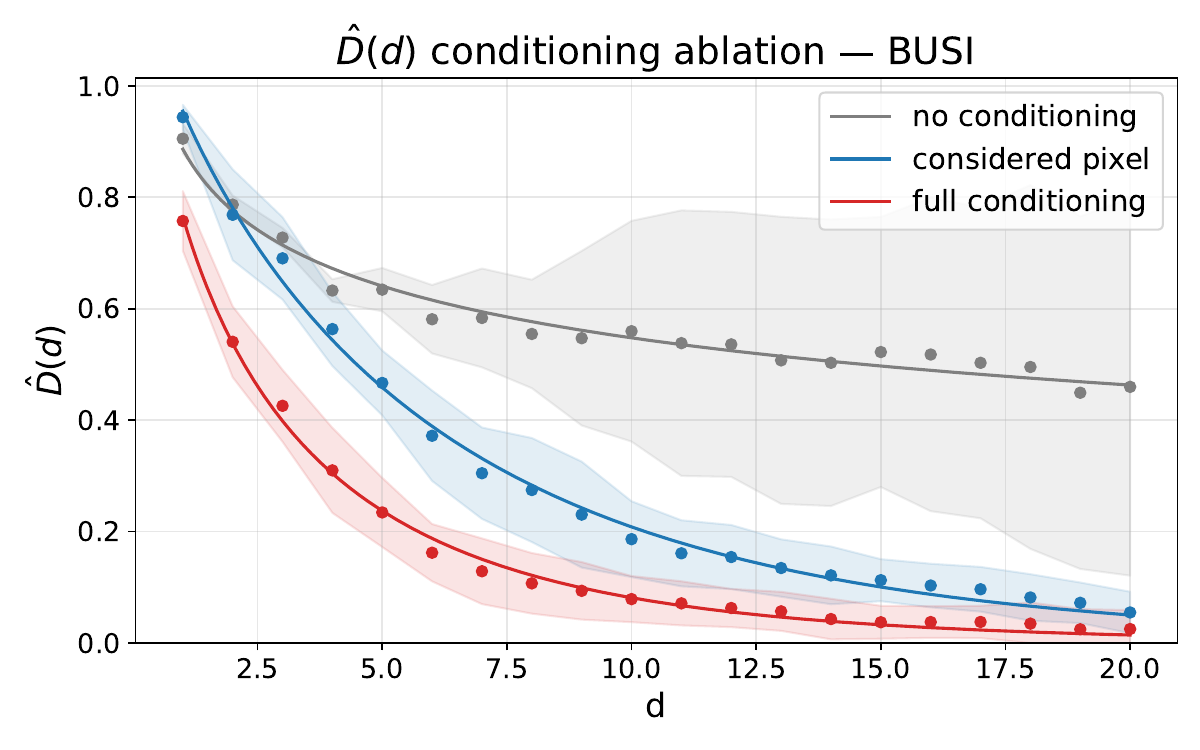}
    \caption{Estimation of $\hat{D}(d)$ with different autoregressive models}
    \label{fig:abst}
\end{figure}

\changes{To assess the assumption empirically, we set up the following experiment.
We train three autoregressive models and use them to approximate the matrix $\hat{D}(d)$ as in Section}~\ref{D_est}.
\changes{Each model is trained with a different binning conditioning.
The first uses no conditioning and learns $\mathbb{P}_\theta(\sigma)$; the second is conditioned on $h(X)_x$, which is the setup used in our main experiments; and the third is conditioned on the full binning pattern, i.e.\ $\mathbb{P}_\theta(\sigma\mid h(X))$.
The results are reported in Fig.}~\ref{fig:abst}.
\changes{The experiments neither reject nor confirm the assumption outright. 
As expected, however, they show that for some datasets, e.g Pet, conditioning on the full binning pattern has no---or only negligible---effect, whereas for others, e.g BUSI, the binning information contained in the neighborhood appears to influence the distribution of the label.
A plausible explanation is that BUSI images often contain small-scale tumors, for which the surrounding pixel context is more informative, whereas the Pet dataset typically exhibits smooth, large-scale segmentation masks.
Relaxing the conditioning dependence assumption, and deriving the corresponding generalization of the bound, is therefore a promising direction for future work.}
\end{document}

%% file: math_commands.tex
\usepackage{amsmath,amsfonts,bm}

\def\eqref#1{equation~\ref{#1}}

\def\1{\bm{1}}

\DeclareMathAlphabet{\mathsfit}{\encodingdefault}{\sfdefault}{m}{sl}
\SetMathAlphabet{\mathsfit}{bold}{\encodingdefault}{\sfdefault}{bx}{n}

\newcommand{\E}{\mathbb{E}}

\newcommand{\Var}{\mathrm{Var}}



%% file: main.bib
@article{dawid1982well,
  title={{The Well-Calibrated Bayesian}},
  author={Dawid, A Philip},
  journal={Journal of the American statistical Association},
  volume={77},
  number={379},
  pages={605--610},
  year={1982},
  publisher={Taylor \& Francis}
}

@article{ovadia2019can,
  title={{Can You Trust Your Model's Uncertainty? Evaluating Predictive Uncertainty Under Dataset Shift}},
  author={Ovadia, Yaniv and Fertig, Emily and Ren, Jie and Nado, Zachary and Sculley, David and Nowozin, Sebastian and Dillon, Joshua and Lakshminarayanan, Balaji and Snoek, Jasper},
  journal={{Advances in Neural Information Processing Systems}},
  volume={32},
  year={2019}
}

@article{kull2019beyond,
  title={Beyond temperature scaling: Obtaining well-calibrated multi-class probabilities with {D}irichlet calibration},
  author={Kull, Meelis and Perello Nieto, Miquel and K{\"a}ngsepp, Markus and Silva Filho, Telmo and Song, Hao and Flach, Peter},
  journal={{Advances in Neural Information Processing Systems}},
  volume={32},
  year={2019}
}

@inproceedings{kull2017beta,
  title={{Beta} calibration: a well-founded and easily implemented improvement on logistic calibration for binary classifiers},
  author={Kull, Meelis and Silva Filho, Telmo and Flach, Peter},
  booktitle={{Artificial Intelligence and Statistics (AISTATS)}},
  pages={623--631},
  year={2017},
  organization={PMLR}
}

@article{platt1999probabilistic,
  title={{Probabilistic Outputs for Support Vector Machines and Comparisons to Regularized Likelihood Methods}},
  author={Platt, John and others},
  journal={Advances in Large Margin Classifiers},
  volume={10},
  number={3},
  pages={61--74},
  year={1999},
  publisher={Cambridge, MA}
}

@inproceedings{zadrozny2001obtaining,
  title={Obtaining calibrated probability estimates from decision trees and naive {B}ayesian classifiers},
  author={Zadrozny, Bianca and Elkan, Charles},
  booktitle={{ICML}},
  pages={2001},
  year={2001}
}

@inproceedings{zadrozny2002transforming,
  title={{Transforming Classifier Scores into Accurate Multiclass Probability Estimates}},
  author={Zadrozny, Bianca and Elkan, Charles},
  booktitle={Proceedings of the eighth ACM SIGKDD International Conference on Knowledge Discovery and Data Mining},
  pages={694--699},
  year={2002}
}

@article{rousseau2025post,
  title={Post hoc calibration of medical segmentation models},
  author={Rousseau, Axel-Jan and Becker, Thijs and Appeltans, Simon and Blaschko, Matthew and Valkenborg, Dirk},
  journal={Discover Applied Sciences},
  volume={7},
  number={3},
  pages={180},
  year={2025},
  publisher={Springer}
}

@inproceedings{ma2021meta,
  title={{Meta-Cal: Well-controlled post-hoc calibration by ranking}},
  author={Ma, Xingchen and Blaschko, Matthew B},
  booktitle={International Conference on Machine Learning},
  pages={7235--7245},
  year={2021},
  organization={PMLR}
}

@inproceedings{ding2021local,
  title={{Local Temperature Scaling for Probability Calibration}},
  author={Ding, Zhipeng and Han, Xu and Liu, Peirong and Niethammer, Marc},
  booktitle={Proceedings of the IEEE/CVF International Conference on Computer Vision},
  pages={6889--6899},
  year={2021}
}

@inproceedings{wang2023calibrating,
  title={{On calibrating semantic segmentation models: Analyses and an algorithm}},
  author={Wang, Dongdong and Gong, Boqing and Wang, Liqiang},
  booktitle={Proceedings of the IEEE/CVF Conference on Computer Vision and Pattern Recognition},
  pages={23652--23662},
  year={2023}
}

@inproceedings{guo2017calibration,
  title={{On Calibration of Modern Neural Networks}},
  author={Guo, Chuan and Pleiss, Geoff and Sun, Yu and Weinberger, Kilian Q},
  booktitle={International Conference on Machine Learning},
  pages={1321--1330},
  year={2017},
  organization={PMLR}
}

@incollection{kuppers2022confidence,
  title={{Confidence Calibration for Object Detection and Segmentation}},
  author={K{\"u}ppers, Fabian and Haselhoff, Anselm and Kronenberger, Jan and Schneider, Jonas},
  booktitle={Deep Neural Networks and Data for Automated Driving: Robustness, Uncertainty Quantification, and Insights Towards Safety},
  pages={225--250},
  year={2022},
  publisher={Springer}
}

@inproceedings{szegedy2016rethinking,
  title={{Rethinking the Inception Architecture for Computer Vision}},
  author={Szegedy, Christian and Vanhoucke, Vincent and Ioffe, Sergey and Shlens, Jon and Wojna, Zbigniew},
  booktitle={Proceedings of the IEEE Conference on Computer Vision and Pattern Recognition},
  pages={2818--2826},
  year={2016}
}

@inproceedings{
zhang2017mixup,
title={{mixup: Beyond Empirical Risk Minimization}},
author={Hongyi Zhang and Moustapha Cisse and Yann N. Dauphin and David Lopez-Paz},
booktitle={International Conference on Learning Representations},
year={2018},
}

@inproceedings{yun2019cutmix,
  title={{CutMix: Regularization strategy to train strong classifiers with localizable features}},
  author={Yun, Sangdoo and Han, Dongyoon and Oh, Seong Joon and Chun, Sanghyuk and Choe, Junsuk and Yoo, Youngjoon},
  booktitle={Proceedings of the IEEE/CVF International Conference on Computer Vision},
  pages={6023--6032},
  year={2019}
}

@article{muller2019does,
  title={{When Does Label Smoothing Help?}},
  author={M{\"u}ller, Rafael and Kornblith, Simon and Hinton, Geoffrey E},
  journal={{Advances in Neural Information Processing Systems}},
  volume={32},
  year={2019}
}

@inproceedings{kumar2018trainable,
  title={{Trainable Calibration Measures For Neural Networks From Kernel Mean Embeddings}},
  author={Kumar, Aviral and Sarawagi, Sunita and Jain, Ujjwal},
  booktitle={International Conference on Machine Learning},
  pages={2805--2814},
  year={2018},
  organization={PMLR}
}

@article{bohdal2023metacalibration,
title={{Meta-Calibration: Learning of Model Calibration Using Differentiable Expected Calibration Error}},
author={Ondrej Bohdal and Yongxin Yang and Timothy Hospedales},
journal={Transactions on Machine Learning Research},
issn={2835-8856},
year={2023},
note={}
}

@article{barfoot2026average,
  title={{Average Calibration Losses for Reliable Uncertainty in Medical Image Segmentation}},
  author={Barfoot, Theodore and Garcia-Peraza-Herrera, Luis C and Akcay, Samet and Glocker, Ben and Vercauteren, Tom},
  journal={IEEE Transactions on Medical Imaging},
  year={2026},
  publisher={IEEE}
}

@inproceedings{barfoot2024average,
  title={Average {C}alibration {E}rror: A {D}ifferentiable {L}oss for {I}mproved {R}eliability in {I}mage {S}egmentation},
  author={Barfoot, Theodore and Garcia Peraza Herrera, Luis C and Glocker, Ben and Vercauteren, Tom},
  booktitle={International Conference on Medical Image Computing and Computer-Assisted Intervention},
  pages={139--149},
  year={2024},
  organization={Springer}
}

@article{mukhoti2020calibrating,
  title={{Calibrating Deep Neural Networks using Focal Loss}},
  author={Mukhoti, Jishnu and Kulharia, Viveka and Sanyal, Amartya and Golodetz, Stuart and Torr, Philip and Dokania, Puneet},
  journal={{Advances in Neural Information Processing Systems}},
  volume={33},
  pages={15288--15299},
  year={2020}
}

@inproceedings{lin2017focal,
  title={{Focal Loss for Dense Object Detection}},
  author={Lin, Tsung-Yi and Goyal, Priya and Girshick, Ross and He, Kaiming and Doll{\'a}r, Piotr},
  booktitle={Proceedings of the IEEE International Conference on Computer Vision},
  pages={2980--2988},
  year={2017}
}

@inproceedings{hebbalaguppe2022stitch,
  title={{A Stitch in Time Saves Nine: A Train-Time Regularizing Loss for Improved Neural Network Calibration}},
  author={Hebbalaguppe, Ramya and Prakash, Jatin and Madan, Neelabh and Arora, Chetan},
  booktitle={Proceedings of the IEEE/CVF Conference on Computer Vision and Pattern Recognition},
  pages={16081--16090},
  year={2022}
}

@article{murugesan2025neighbor,
  title={{ Neighbor-Aware Calibration of Segmentation Networks with Penalty-Based Constraints }},
  author={Murugesan, Balamurali and Vasudeva, Sukesh Adiga and Liu, Bingyuan and Lombaert, Herve and Ayed, Ismail Ben and Dolz, Jose},
  journal={Medical Image Analysis},
  volume={101},
  pages={103501},
  year={2025},
  publisher={Elsevier}
}

@inproceedings{murugesan2023trust,
  title={{ Trust your neighbours: Penalty-based constraints for model calibration }},
  author={Murugesan, Balamurali and Adiga Vasudeva, Sukesh and Liu, Bingyuan and Lombaert, Herve and Ben Ayed, Ismail and Dolz, Jose},
  booktitle={International Conference on Medical Image Computing and Computer-Assisted Intervention},
  pages={572--581},
  year={2023},
  organization={Springer}
}

@inproceedings{gal2016dropout,
  title={{Dropout as a Bayesian Approximation: Representing Model Uncertainty in Deep Learning}},
  author={Gal, Yarin and Ghahramani, Zoubin},
  booktitle={International Conference on Machine Learning},
  pages={1050--1059},
  year={2016},
  organization={PMLR}
}

@article{kendall2017uncertainties,
  title={{What Uncertainties Do We Need in Bayesian Deep Learning for Computer Vision?}},
  author={Kendall, Alex and Gal, Yarin},
  journal={{Advances in Neural Information Processing Systems}},
  volume={30},
  year={2017}
}

@article{lakshminarayanan2017simple,
  title={{Simple and Scalable Predictive Uncertainty Estimation using Deep Ensembles}},
  author={Lakshminarayanan, Balaji and Pritzel, Alexander and Blundell, Charles},
  journal={{Advances in Neural Information Processing Systems}},
  volume={30},
  year={2017}
}

@article{mehrtash2020confidence,
  title={{Confidence Calibration and Predictive Uncertainty Estimation for Deep Medical Image Segmentation}},
  author={Mehrtash, Alireza and Wells, William M and Tempany, Clare M and Abolmaesumi, Purang and Kapur, Tina},
  journal={IEEE Transactions on Medical Imaging},
  volume={39},
  number={12},
  pages={3868--3878},
  year={2020},
  publisher={IEEE}
}

@book{vovk2005algorithmic,
	title={{Algorithmic Learning in a Random World}},
	author={Vovk, Vladimir and Gammerman, Alexander and Shafer, Glenn},
	year={2005},
	publisher={Springer}
}

@article{liu2025spatial,
  title={{Spatial-Aware Conformal Prediction for Trustworthy Hyperspectral Image Classification}},
  author={Liu, Kangdao and Sun, Tianhao and Zeng, Hao and Zhang, Yongshan and Pun, Chi-Man and Vong, Chi-Man},
  journal={IEEE Transactions on Circuits and Systems for Video Technology},
  year={2025},
  publisher={IEEE}
}

@inproceedings{mossina2025conformal,
  title={{Conformal Prediction for Image Segmentation Using Morphological Prediction Sets}},
  author={Mossina, Luca and Friedrich, Corentin},
  booktitle={International Conference on Medical Image Computing and Computer-Assisted Intervention},
  pages={78--88},
  year={2025},
  organization={Springer}
}

@inproceedings{bereska2025sacp,
  title={{SACP: Spatially-Adaptive Conformal Prediction in Uncertainty Quantification of Medical Image Segmentation}},
  author={Bereska, Jacqueline Isabel and Karimi, Hamed and Samavi, Reza},
  booktitle={Medical Imaging with Deep Learning},
  year={2025}
}

@inproceedings{
teng2023predictive,
title={{Predictive Inference with Feature Conformal Prediction}},
author={Jiaye Teng and Chuan Wen and Dinghuai Zhang and Yoshua Bengio and Yang Gao and Yang Yuan},
booktitle={The Eleventh International Conference on Learning Representations },
year={2023}
}

@inproceedings{
viti2026consign,
title={{CONSIGN: Conformal Segmentation Informed by Spatial Groupings via Decomposition}},
author={Bruno Viti and Elias Karabelas and Martin Holler},
booktitle={The Fourteenth International Conference on Learning Representations},
year={2026}
}

@inproceedings{brunekreef2024kandinsky,
  title={{Kandinsky Conformal Prediction: Efficient Calibration of Image Segmentation Algorithms}},
  author={Brunekreef, Joren and Marcus, Eric and Sheombarsing, Ray and Sonke, Jan-Jakob and Teuwen, Jonas},
  booktitle={Proceedings of the IEEE/CVF Conference on Computer Vision and Pattern Recognition},
  pages={4135--4143},
  year={2024}
}

@article{
xi2025does,
title={{Does confidence calibration improve conformal prediction?}},
author={HuaJun Xi and Jianguo Huang and Kangdao Liu and Lei Feng and Hongxin Wei},
journal={Transactions on Machine Learning Research},
issn={2835-8856},
year={2025},
note={}
}

@article{hoeff,
 ISSN = {01621459, 1537274X},
 author = {Wassily Hoeffding},
 journal = {Journal of the American Statistical Association},
 number = {301},
 pages = {13--30},
 publisher = {[American Statistical Association, Taylor & Francis, Ltd.]},
 title = {{Probability Inequalities for Sums of Bounded Random Variables}},
 urldate = {2026-03-13},
 volume = {58},
 year = {1963}
}

@book{Massart,
    author = {Boucheron, Stéphane and Lugosi, Gábor and Massart, Pascal},
    title = {{Concentration Inequalities: A Nonasymptotic Theory of Independence}},
    publisher = {Oxford University Press},
    year = {2013},
    month = {02},
    isbn = {9780199535255},
}

@article{mcdiarmid1989method,
  title={On the method of bounded differences},
  author={McDiarmid, Colin and others},
  journal={Surveys in combinatorics},
  volume={141},
  number={1},
  pages={148--188},
  year={1989},
  publisher={Norwich}
}

@article{chazottes2007,
  title={Concentration inequalities for random fields via coupling},
  author={Chazottes, J-R and Collet, Pierre and K{\"u}lske, Christof and Redig, Frank},
  journal={Probability Theory and Related Fields},
  volume={137},
  number={1},
  pages={201--225},
  year={2007},
  publisher={Springer}
}

@article{kontorovich2008concentration,
  title={Concentration inequalities for dependent random variables via the martingale method},
  author={Kontorovich, Leonid and Ramanan, Kavita},
  year={2008}
}

@article{azuma1967weighted,
  title={Weighted sums of certain dependent random variables},
  author={Azuma, Kazuoki},
  journal={Tohoku Mathematical Journal, Second Series},
  volume={19},
  number={3},
  pages={357--367},
  year={1967},
  publisher={Mathematical Institute, Tohoku University}
}

@article{freedman1975tail,
  title={On tail probabilities for martingales},
  author={Freedman, David A},
  journal={the Annals of Probability},
  pages={100--118},
  year={1975},
  publisher={JSTOR}
}

@article{yu1994rates,
  title={Rates of convergence for empirical processes of stationary mixing sequences},
  author={Yu, Bin},
  journal={The Annals of Probability},
  pages={94--116},
  year={1994},
  publisher={JSTOR}
}

@article{mohri2010stability,
  title={{Stability Bounds for Stationary} $\varphi$-mixing {and} $\beta$-mixing {Processes.}},
  author={Mohri, Mehryar and Rostamizadeh, Afshin},
  journal={Journal of Machine Learning Research},
  volume={11},
  number={2},
  year={2010}
}

@article{garnier2022,
  author  = {Garnier, Josselin and Langhendries, Xavier},
  title   = {{Concentration Inequalities for Non-causal Random Fields}},
  journal = {Electronic Journal of Statistics},
  year    = {2022},
  volume  = {16},
  pages   = {1--44}
}

@article{chazottes2026finitary,
  author  = {Chazottes, Jean-Ren{\'e} and others},
  title   = {{Finitary Coding and Gaussian Concentration for Random Fields}},
  journal = {arXiv preprint arXiv:2602.14618},
  year    = {2026}
}

@article{marton1998,
  author  = {Marton, Katalin},
  title   = {{Measure Concentration for a Class of Random Processes}},
  journal = {Probability Theory and Related Fields},
  year    = {1998},
  volume  = {110},
  pages   = {427--439}
}

@article{marton2003,
  author  = {Marton, Katalin},
  title   = {{Measure Concentration and Strong Mixing}},
  journal = {Studia Scientiarum Mathematicarum Hungarica},
  year    = {2003},
  volume  = {40},
  pages   = {95--113}
}

@article{samson2000,
  author  = {Samson, Paul-Marie},
  title   = {{Concentration of Measure Inequalities for Markov Chains and} $\Phi$-mixing {Processes}},
  journal = {The Annals of Probability},
  year    = {2000},
  volume  = {28},
  pages   = {416--461}
}

@article{zoltan2012markov,
  title={Markov random fields in image segmentation},
  author={Zoltan, Kato and Josiane, Zerubia},
  journal={Foundations and Trends in Signal Processing},
  volume={5},
  number={1-2},
  pages={1--155},
  year={2012},
  publisher={Emerald Publishing Limited}
}

@book{li2009markov,
  author    = {Li, Stan Z.},
  title     = {{Markov Random Field Modeling in Image Analysis}},
  publisher = {Springer},
  year      = {2009},
  edition   = {3rd}
}

@book{lindvall2002lectures,
  title={Lectures on the coupling method},
  author={Lindvall, Torgny},
  year={2002},
  publisher={Courier Corporation}
}

@inproceedings{gupta2021distribution,
  title={{Distribution-free calibration guarantees for histogram binning without sample splitting}},
  author={Gupta, Chirag and Ramdas, Aaditya},
  booktitle={International Conference on Machine Learning},
  pages={3942--3952},
  year={2021},
  organization={PMLR}
}

@article{gupta2020distribution,
  title={{Distribution-free binary classification: prediction sets, confidence intervals and calibration}},
  author={Gupta, Chirag and Podkopaev, Aleksandr and Ramdas, Aaditya},
  journal={{Advances in Neural Information Processing Systems}},
  volume={33},
  pages={3711--3723},
  year={2020}
}

@inproceedings{Pogorelov:2017:KMI:3083187.3083212,
  title = {{KVASIR: A Multi-Class Image Dataset for Computer Aided Gastrointestinal Disease Detection}},
  author = {
     Pogorelov, Konstantin and Randel, Kristin Ranheim and Griwodz, Carsten and
     Eskeland, Sigrun Losada and de Lange, Thomas and Johansen, Dag and
     Spampinato, Concetto and Dang-Nguyen, Duc-Tien and Lux, Mathias and
     Schmidt, Peter Thelin and Riegler, Michael and Halvorsen, P{\aa}l
  },
  booktitle = {Proceedings of the 8th ACM on Multimedia Systems Conference},
  series = {MMSys'17},
  year = {2017},
  isbn = {978-1-4503-5002-0},
  location = {Taipei, Taiwan},
  pages = {164--169},
  numpages = {6},
  doi = {10.1145/3083187.3083212},
  acmid = {3083212},
  publisher = {ACM},
  address = {New York, NY, USA},
}

@article{GONG2023106389,
title = {Thyroid region prior guided attention for ultrasound segmentation of thyroid nodules},
journal = {Computers in Biology and Medicine},
volume = {155},
pages = {106389},
year = {2023},
issn = {0010-4825},
author = {Haifan Gong and Jiaxin Chen and Guanqi Chen and Haofeng Li and Guanbin Li and Fei Chen}
}

@article{ign2022flair1,
  doi = {10.13140/RG.2.2.30183.73128/1},
  author = {Garioud, Anatol and Peillet, Stéphane and Bookjans, Eva and Giordano, Sébastien and Wattrelos, Boris},
  title = {{FLAIR} {\#}1: semantic segmentation and domain adaptation dataset},
  publisher = {arXiv},
  year = {2022}
}

@article{gutman2016skin,
  title={Skin lesion analysis toward melanoma detection: A challenge at the international symposium on biomedical imaging ({ISBI}) 2016, hosted by the international skin imaging collaboration ({ISIC})},
  author={Gutman, David and Codella, Noel CF and Celebi, Emre and Helba, Brian and Marchetti, Michael and Mishra, Nabin and Halpern, Allan},
  journal={arXiv preprint arXiv:1605.01397},
  year={2016}
}

@inproceedings{Cordts2016Cityscapes,
title={{The Cityscapes Dataset for Semantic Urban Scene Understanding}},
author={Cordts, Marius and Omran, Mohamed and Ramos, Sebastian and Rehfeld, Timo and Enzweiler, Markus and Benenson, Rodrigo and Franke, Uwe and Roth, Stefan and Schiele, Bernt},
booktitle={Proc. of the IEEE Conference on Computer Vision and Pattern Recognition (CVPR)},
year={2016}
}

@inproceedings{Cordts2015Cvprw,
title={{The Cityscapes Dataset}},
author={Cordts, Marius and Omran, Mohamed and Ramos, Sebastian and Scharw{\"a}chter, Timo and Enzweiler, Markus and Benenson, Rodrigo and Franke, Uwe and Roth, Stefan and Schiele, Bernt},
booktitle={CVPR Workshop on The Future of Datasets in Vision},
year={2015}
}

@inproceedings{lin2014microsoft,
  title={{Microsoft COCO: Common Objects in Context}},
  author={Lin, Tsung-Yi and Maire, Michael and Belongie, Serge and Hays, James and Perona, Pietro and Ramanan, Deva and Doll{\'a}r, Piotr and Zitnick, C Lawrence},
  booktitle={Computer Vision--ECCV 2014: 13th European conference, zurich, Switzerland, September 6-12, 2014, proceedings, part v 13},
  pages={740--755},
  year={2014},
  organization={Springer}
}

@InProceedings{parkhi12a,
  author       = "Omkar M. Parkhi and Andrea Vedaldi and Andrew Zisserman and C. V. Jawahar",
  title        = "{Cats and Dogs}",
  booktitle    = "IEEE Conference on Computer Vision and Pattern Recognition",
  year         = "2012",
}

@inproceedings{ronneberger2015u,
  title={{U-Net: Convolutional Networks for Biomedical Image Segmentation}},
  author={Ronneberger, Olaf and Fischer, Philipp and Brox, Thomas},
  booktitle={International Conference on Medical Image Computing and Computer-Assisted Intervention},
  pages={234--241},
  year={2015},
  organization={Springer}
}

@article{tasnim2024cam,
  title={{CAM-QUS} guided self-tuning modular {CNNs} with multi-loss functions for fully automated breast lesion classification in ultrasound images},
  author={Tasnim, Jarin and Hasan, Md Kamrul},
  journal={Physics in Medicine \& Biology},
  volume={69},
  number={1},
  pages={015018},
  year={2024},
  publisher={IOP Publishing}
}

@article{al2020dataset,
  title={Dataset of breast ultrasound images},
  author={Al-Dhabyani, Walid and Gomaa, Mohammed and Khaled, Hussien and Fahmy, Aly},
  journal={Data in brief},
  volume={28},
  pages={104863},
  year={2020},
  publisher={Elsevier}
}

@article{chen2019rethinking,
  title={{Rethinking Atrous Convolution for Semantic Image Segmentation}},
  author={Chen, Liang-Chieh and Papandreou, George and Schroff, Florian and Adam, Hartwig},
  journal={arXiv preprint arXiv:1706.05587},
  year={2017}
}

@inproceedings{chen2018encoder,
  title={{Encoder-Decoder with Atrous Separable Convolution for Semantic Image Segmentation}},
  author={Chen, Liang-Chieh and Zhu, Yukun and Papandreou, George and Schroff, Florian and Adam, Hartwig},
  booktitle={Proceedings of the European Conference on Computer Vision (ECCV)},
  pages={801--818},
  year={2018}
}

@article{mohamed2020monte,
  title={{Monte Carlo Gradient Estimation in Machine Learning}},
  author={Mohamed, Shakir and Rosca, Mihaela and Figurnov, Michael and Mnih, Andriy},
  journal={Journal of Machine Learning Research},
  volume={21},
  number={132},
  pages={1--62},
  year={2020}
}

@book{glasserman2004monte,
  title={{Monte Carlo Methods in Financial Engineering}},
  author={Glasserman, Paul},
  volume={53},
  year={2004},
  publisher={Springer}
}

@inproceedings{van2016pixel,
  title={{Pixel Recurrent Neural Networks}},
  author={Van Den Oord, A{\"a}ron and Kalchbrenner, Nal and Kavukcuoglu, Koray},
  booktitle={International Conference on Machine Learning},
  pages={1747--1756},
  year={2016},
  organization={PMLR}
}

@article{van2016conditional,
  title={{Conditional Image Generation with PixelCNN Decoders}},
  author={Van Den Oord, Aaron and Kalchbrenner, Nal and Espeholt, Lasse and Vinyals, Oriol and Graves, Alex and others},
  journal={{Advances in Neural Information Processing Systems}},
  volume={29},
  year={2016}
}
